\documentclass[3p,times]{elsarticle}

\usepackage{ecrc}

\volume{00}

\firstpage{1}

\journalname{International Journal of Approximate Reasoning}

\runauth{Jakubův, Janota, Piepenbrock, Urban}

\jid{procs}

\CopyrightLine{2011}{Published by Elsevier Ltd.}

\usepackage{amssymb}
\usepackage[figuresright]{rotating}

\usepackage{tikz}
\usetikzlibrary{arrows,shapes.geometric}
\usetikzlibrary{shapes.symbols}

\usepackage{latexsym}
\usepackage{amsmath}
\usepackage{amsthm}
\usepackage{booktabs}
\usepackage{enumitem}
\usepackage{graphicx}
\usepackage{color}
\usepackage{url}
\usepackage{bold-extra}

\usepackage{numprint}
\npthousandsep{\thinspace}
\npdecimalsign{.}
\usepackage[binary-units=true,detect-weight=true]{siunitx}
\usepackage{xspace}

\newcommand{\BibTeX}{B\kern-.05em{\sc i\kern-.025em b}\kern-.08em\TeX}

\newcommand*{\grk}[1]{\ensuremath{\mathsf{grk}_{#1}}}
\newcommand*{\casc}[1]{\ensuremath{\mathsf{casc}_{#1}}}

\newcommand*{\sparrow}{\grk{1}}
\newcommand*{\buzzard}{\grk{2}}
\newcommand*{\chickadee}{\grk{3}}

\newcommand*{\Mmerge}{\ensuremath{\mathsf{merge}}\xspace}
\newcommand*{\Msparrow}{\ensuremath{\mathsf{single}}\xspace}

\newcommand*{\feature}[2]{\ensuremath{\mathsf{#1}^\mathsf{#2}}}

\usepackage[textsize=scriptsize]{todonotes} %
\usepackage[none]{hyphenat}

\begin{document}

\begin{frontmatter}

\dochead{}

\title{Machine Learning for Quantifier Selection in cvc5\tnoteref{ack}}

\author[A]{Jan Jakub\r{u}v\corref{auth:jj}}
\ead{jakubuv@gmail.com}
\author[A]{Mikol\'{a}\v{s} Janota}
\author[A,B]{Jelle Piepenbrock}
\author[A]{Josef Urban}
\cortext[auth:jj]{corresponding author}

\address[A]{Czech Institute of Informatics, Robotics and Cybernetics, Czech Technical University in Prague, Czechia}
\address[B]{Institute for Computing and Information Sciences, Radboud University, Nijmegen, the Netherlands}

\tnotetext[ack]{Supported by the Czech MEYS under the ERC CZ project no.\ LL1902 \emph{POSTMAN} (JJ, MJ, JP, JU),
Amazon Research Awards (JU), EU ICT-48 2020 project no.~952215 \emph{TAILOR} (JU), 
ERC PoC grant no.~101156734 \emph{FormalWeb3} (JJ), and the Czech Science Foundation
grant 25-17929X.
The research was co-funded by the European Union under the project \emph{ROBOPROX}
(reg.~no.~CZ.02.01.01/00/22\_008/0004590).
This article is part of the \emph{RICAIP} project that has received funding
from the European Union's Horizon~2020 research and innovation programme under
grant agreement No~857306.}

\begin{abstract}
In this work we considerably improve   the real-time performance of
  state-of-the-art SMT solving on first-order quantified problems by
efficient machine learning guidance of quantifier selection.
Quantifiers represent a significant challenge for SMT and are
technically a source of undecidability.
In our approach, we train an efficient machine learning
model that informs the solver which quantifiers should be instantiated
and which not.
Each quantifier may be instantiated multiple times and the set of 
the currently
active quantifiers changes as the solving progresses.
Therefore, we invoke the ML predictor many times,
during the whole run of the solver. %
To make this efficient, we use fast ML models based on gradient boosted decision trees.
We integrate our approach into the state-of-the-art cvc5 SMT solver and show a considerable
increase of the system's holdout-set performance after training it on
large sets of first-order problems. The method is tested in several ways, using both single-strategy and portfolio
approaches. The evaluation is done on two large formal verification corpora: first-order problems created from the Mizar
Mathematical Library, and first-order problems created from the HOL4 standard library.
\end{abstract}


\end{frontmatter}

\section{Introduction}\label{sec:intro}

The use of machine learning methods in various fields of automated reasoning is
an emerging research topic with successful applications~\cite{DBLP:conf/birthday/BlaauwbroekCGJK24,JakubuvU17a,DBLP:conf/frocos/Suda21}.
Machine learning methods were previously integrated into various provers,
solvers, and related systems~\cite{JakubuvCGKOP00U23,GoertzelJKOPU22}.
Here, we focus on the task of \emph{quantifier
selection} within the state-of-the-art \emph{Satisfiability Modulo Theory}
(SMT) solver, cvc5, where various instantiation methods are applied to
quantified formulas to refine problem representation. 
While cvc5's instantiation methods generate instantiations for all applicable
quantified formulas, we aim to filter out the formulas considered by the
instantiation module using a machine-learned predictor trained on previously
proved similar problems.
We refer to this process as \emph{quantifier selection} for simplicity.

In contrast to previous research efforts that concentrated on fine-grained
control of the solver at the 
\emph{term level}~\cite{bandits,janota-sat22}, we simplify the problem by shifting
our focus to controlling the solver at the \emph{quantifier level}.
This simplification allows us to employ our method
basically with any quantifier instantiation method supported by cvc5. Our
methods also improve upon a related method of \emph{offline premise selection}~\cite{abs-1108-3446,GoertzelJKOPU22},
where initial quantified assumptions are filtered \emph{only once} before launching the solver.
Such filtering can cause irrecoverable mistakes. For example, a \emph{deletion
abstraction}~\cite{DBLP:conf/cade/HernandezK18} may produce an overly weak
theory when a necessary assumption is deleted. 
In contrast, the method presented here ensures that every quantified formula is
considered for instantiation with non-zero probability.
This yields a more complex, probabilistically guided framework implementing deletion and instantiation abstractions in the framework proposed in~\cite{DBLP:conf/cade/HernandezK18}.
As we employ a
highly effective version of \emph{gradient boosting decision trees} with %
efficiently computable \emph{symbol-based features}, %
our implementation produces
only a minimal overhead over the standard cvc5 run.

The simplicity of our approach allows us to apply our methods extensively and
train on a large number of problems created from two large formal verification corpora: the \emph{Mizar Mathematical Library}
(MML)~\cite{BancerekBGKMNP18}, and the HOL4~\cite{SlindN08,GordonMelham93} standard library. Our methods also exhibit remarkable qualities of cross-strategy \emph{model
transfer}, where a model trained on samples from one strategy improves the
performance of virtually any other strategy. This allows us to improve the
state-of-the-art performance of cvc5 on MML by more than 20\% in both the
\emph{single strategy} and \emph{portfolio} scenarios. Our best machine-learned
strategy, alone, outperforms the state-of-the-art cvc5's portfolio from the
latest CASC competition by more than 10\%.

The rest of this paper is structured as follows. Section~\ref{sec:backgound}
provides a basic introduction to SMT solving. Section~\ref{sec:cvc5ml}
describes the core of our method to equip cvc5 with machine learning.
Section~\ref{sec:exp} proceeds with straightforward and reproducible
experiments to gauge the benefit of our method. Section~\ref{sec:exp-extra}
investigates the issue of cross-strategy knowledge transfer and provides
further insights into the functioning of the trained models through feature
importance analysis. Finally, we discuss several hard Mizar problems proved by
the new methods in Section~\ref{sec:newprobs}. We conclude in
Section~\ref{sec:conclusion}.

This journal publication significantly extends our previous ECAI'24
paper~\cite{JakubuvJPU24} by expanding the feature representation used for
constructing ML-enhanced strategy portfolios. While our earlier approach relied
solely on a bag-of-words model, we now also incorporate efficiently computed structured \emph{path features} that
capture both vertical and horizontal cuts of the formula syntax tree.
Additionally, we embed symbol type information into the feature vectors while
maintaining an anonymous representation that remains independent of specific
symbol names. These enhancements provide the system with a richer understanding
of SMT quantified formulas, leading to improved predictive performance.
Furthermore, we evaluate our approach on a completely new benchmark dataset  --- the HOL4 standard library ---
demonstrating the effectiveness of the methods across a broader range of problems and
achieving strong empirical results.
The first-time presented results are mainly concentrated in
Section~\ref{sec:ml-vectors}, Section~\ref{sec:exp},
Section~\ref{sec:exp-extra}, and Section~\ref{sec:newprobs}. Although the
experiments in this journal version partially overlap with the ECAI paper, the
experiments here were conducted anew using a compatible implementation.

\section{Background: SMT solving}\label{sec:backgound}

\begin{figure}[t]
  \centering
  \usetikzlibrary{positioning}
  \begin{tikzpicture}[font=\sffamily,thick,xscale=1.55,yscale=0.8,%
    module/.style={draw=gray!40!black,fill=blue!10,rounded corners=2pt},
    db/.style={cloud,draw=red,thick,fill=red!20,minimum height=1em,aspect=3}
    ]
    \draw[draw=gray] (-2.2,-0.01) rectangle (1.6,1.8);
    \draw[very thick,->] (-3.0,.8) -- node[align=center,above] {input\\formula} (-2.2,.8);

    \node[module] (DB) at (1,1.35) {Term DB};
    \node[above right,draw=gray,align=center,fill=green!10] at (-2.2,1.8) {SMT solver};
    \node[right,module,minimum width=2.2cm] (G) at (-1.25,.35) {Ground solver};
    \node[right,module,minimum width=2.2cm] (I) at (-1.25,1.35) {Instantiation};

    \draw[->,>=stealth] (I.east) to[bend left=60] node[near end,right]{\footnotesize instantiations}  (G.east);
    \draw[<->,>=stealth] (I.north east) to[bend left=10] (DB.north west);
    \draw[->,>=stealth] (G.west) to[bend left=60]
    node[align=center,left]{\footnotesize ground\\ \footnotesize model} (I.west);

    \node[right,inner sep=0pt] (s) at (1.8,1.6) {sat};
    \node[right,inner sep=0pt] (u) at (1.8,1.1) {unsat};
    \node[align=left,right,inner sep=0pt] at (1.7,.3) {or\\infinite};

    \draw[very thick,->,>=stealth] (1.6,1.35) to (s.west);
    \draw[very thick,->,>=stealth] (1.6,1.35) to (u.west);
  \end{tikzpicture}
  \caption{Schema of quantifier instantiation in SMT, adapted from~\cite{janota2021fair}.}%
  \label{fig:schematic}
\end{figure}
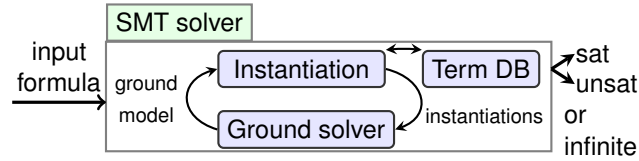

To solve formulas \emph{without} quantifiers (aka \emph{ground} formulas), SMT
solvers combine SAT solving with theory solving~\cite{barrett-faia09}. The SAT
solver handles the Boolean structure of the formula, while theory solvers reason
about concrete theories such as the theory of reals and the theory of integers, etc. Problems in
FOL do not contain such theories explicitly and therefore, to reason about FOL,
an SMT solver needs to only support \emph{uninterpreted functions} and \emph{equality}
(combination known as \texttt{EUF}). For reading this paper, it is not
essential to understand how exactly the SMT solvers solve ground formulas;
we refer interested readers to the relevant literature~\cite{DetlefsNS05,cvc5,barrett-faia09,handbook-clarke}.

Formulas \emph{with} quantifiers represent a significant challenge for
SMT\@. In general, SMT solvers use
\emph{instantiations}---unless they deal with decidable quantified
theories~\cite{Davenport,bjorner-janota-lpar15,ReynoldsKK17}.
Instantiations are done with the goal of achieving a contradiction.
This style of reasoning can be seen as a direct application of the Herbrand's theorem.
For example, for $(\forall x:\mathbb{R}.\,x>0)$ instantiating $x$ with the value \numprint{0}
yields $0>0$, which immediately gives a contradiction (in the theory of reals).

Input formulas do not need to be in prenex form, which effectively means
that the solver may activate only some of the quantified subformulas.
For this purpose, a subformula
$(\forall x_1\dots x_n\,\phi)$ is seen as a generator of lemmas of the form $(\forall
x_1\dots x_n\,\phi)\rightarrow\phi[x_1/t_1,\dots,x_n/t_n]$,
with $t_i$ ground terms. For example,
$\forall x\, R(f(x), c)$ may be instantiated as $(\forall x\, R(f(x), c))\rightarrow
R(f(c),c)$. Existential quantifiers are removed by skolemization.

The solving process alternates between a \emph{ground solver}
and an \emph{instantiation module} (Figure~\ref{fig:schematic}), where the
ground solver perceives quantifiers as opaque propositions. After identifying a
model for the ground part, control shifts to the \emph{instantiation
module}. This module generates new instances of the quantified subformulas that are
currently meant to hold. A new instance is added to the ground part of the formula, thus making it stronger.
The process stops if the ground part becomes unsatisfiable, if ever (model-based quantifier instantiation can also lead to satisfiable answers~\cite{ge-moura-cav09}).

The cvc5 solver implements several instantiation methods.
Some of those can be seen as syntax-driven approaches, \emph{e-matching}~\cite{DetlefsNS05} or syntax-guided instantiation~\cite{niemetz-tacas21}.
Other methods are semantics-driven such as \emph{model-based}~\cite{ge-moura-cav09,reynolds-cade13} or \emph{conflict-based}~\cite{reynolds-fmcad14}.
A straightforward but complete method for FOL is \textit{enumerative
instantiation}~\cite{janota2021fair,ReynoldsBF18} which exhaustively generates
all possible instantiations.
Both e-matching and
enumerative instantiation require the solver to maintain a database of ground
terms. This database grows as new instantiations are performed. In the case of
enumerative instantiation, the terms are selected systematically going from the
oldest to the newest. E-matching tries to instantiate in a way to match an
existing term. 
We dedicate Sections~\ref{sec:enum} and~\ref{sec:ematch} to describing these
two prominent methods because they are the most effective for quantifier
instantiation, and we anticipate that they will derive the greatest benefit
from machine learning.

All these techniques target the quantifiers individually, i.e., the instantiation
of one quantifier does not directly influence the instantiation of another one.
The default implementation of cvc5 is to iterate over all currently-active
quantifiers and instantiate them, one by one. If such instantiation step is
superfluous, the generated lemma not only burdens the underlying SAT solver,
but it also pollutes the considered set of terms with all the lemma's subterms.

\subsection{Enumerative instantiation}\label{sec:enum}
This \emph{enumerative instantiation} mode enumerates all possible tuples of
terms from the term database that can be used to instantiate a quantified
expression~\cite{ReynoldsBF18,janota2021fair}. Given an ordering of the
suitable ground terms for each quantifier, the strategy enumerates tuples by
starting at the tuple that contains the first term in each quantifier-term
ordering, and moves further into the orderings by incrementing the indices that
retrieve further terms in the quantifier-term index.

The particular criterion for term ordering that is used in enumerative
instantiation is the \emph{age} of the ground terms, with terms with higher age
being preferred.
That is, ground terms that were in the original problem have the
highest priority for being tried. As an example of the logic of the
procedure, imagine that there is a ground part  $\{p(c)\}$ and a quantified
expression $\forall x.\,q(f(x))$. In this case, the ground term $c$ is
available at the start of the procedure and therefore will be tried first. This
creates a ground lemma that is a consequence of the quantified expression,
$q(f(c))$. The solver would now recognize that $f(c)$ is also an available
ground term, insert it into the term database and it would be used in the next
round of instantiations.
With nested terms, many new ground terms may be created by a single instantiation step.

\subsection{E-matching}\label{sec:ematch}
The \emph{e-matching} instantiation procedure searches for instantiations that
match some already available ground term~\cite{DetlefsNS05}, taking term
equality into account.
Of course, at any point and especially late in the procedure when
many terms have been generated, there can be many possible ways to create such
matching terms. The e-matching instantiation process is therefore focused using
\emph{triggers}, which are user-supplied or heuristically generated patterns. 

We show the following example for a more concrete perspective. In the example,
we assume that a theory module that can evaluate integer arithmetic statements
is available. Given the following ground facts $\{p(a), a=f(24)\}$ and the
quantified expression  $\forall x.\,\lnot p(f(x))\lor x<0$. A trigger in the
form $p(f(x))$ leads to $x$ being instantiated with the ground term
\numprint{24} as there is an existing ground term $p(f(24))$, when taking equality into
account. Instantiating with \numprint{24} generates the consequent lemma $\lnot
p(f(24))\lor 24<0$. This lemma contradicts the other ground facts
(as the theory of integer arithmetic knows that in fact $24 > 0$), and
therefore the solver stops and reports that a contradiction (unsat) was
found.
An automated trigger generation method is implemented in the e-matching module
of cvc5, and its behavior can be influenced through various options provided by
the user.

\section{Machine-learned quantifier selection}%
\label{sec:cvc5ml}

In this section, we detail our quantifier selection method and its
implementation in the cvc5 solver, covering the process from 
extraction of training examples to  model training, and integration.

\subsection{Instantiation modules}%
\label{sec:ml-modules}

The instantiation methods supported by cvc5 are implemented through various
\emph{instantiation modules} that share a common interface.  An
instantiation module is invoked via its \emph{check} method to refine
information about quantified formulas by introducing appropriate formula instances.
Each module is provided with information about currently asserted quantified
formulas, and it selects formulas and generates their instances in accordance
with the implemented instantiation method. Subsequently, control is returned to
the ground solver.

Effective quantifier and instance selection can significantly enhance
performance, and different instantiation methods offer grounds for various
possible applications of machine learning methods at a method-specific 
level~\cite{bandits,janota-sat22}.
However, we propose a generic method for \emph{quantifier selection} by
limiting the formulas visible to the modules. As all instantiation
modules iterate over available quantified formulas and process them one by one,
we can seamlessly integrate a quantifier selector into any module and simply
skip the processing of undesirable quantifiers. To predict the quality of
quantifiers, we utilize an efficient implementation of \emph{decision tree
ensembles} (LightGBM~\cite{LightGBM}) that enables easy and fast integration with cvc5.
Decision tree models can be trained to classify quantified formulas as
\emph{positive} or \emph{negative} based on provided training examples.
The trained model can be employed within an instantiation module to skip
the processing of negative quantifiers.

\subsection{Feature vectors and path features}%
\label{sec:ml-vectors}

To use decision trees for classifying quantified formulas in cvc5, we need to
represent SMT formulas by numeric feature vectors. SMT formulas within cvc5
are represented using directed acyclic graphs with shared node representation.
The nodes of the graph are labeled with symbols representing logical
connectives, quantifiers, variables, and interpreted/uninterpreted theory
symbols. Each symbol falls into one of finitely many \emph{kinds}, designating
theory-specific (built-in) symbols. Different kinds exist for logical
connectives ($\lnot$, $\land$, $\lor$, etc.), quantifiers ($\forall$,
$\exists$), interpreted symbols ($+$, $<$, \numprint{0}, etc.), and variables. There is
only one kind for all uninterpreted functions, including uninterpreted
constants. We utilize cvc5's kinds as \emph{bag-of-words} features,
representing the quantified formula $q$ with the counts of symbols 
of each kind $k$. 
In particular, we establish an enumeration of the kinds given by internal cvc5
kind IDs (unsigned integers), and the quantifier $q$ is represented by the
bag-of-words vector $\alpha_q$, where $\alpha_q[i]$ denotes the number of
symbols of kind $i$ in $q$.  More than \numprint{300} different kinds are
defined in cvc5.

In addition to the simple bag-of-words features introduced in the ECAI'24
paper~\cite{JakubuvJPU24}, we implement \emph{path features} to enhance the
expressiveness of feature vectors. These path features are inspired by similar
concepts in the ENIGMA system~\cite{JakubuvU18,JakubuvCOP0U20}, where
\emph{vertical} and \emph{horizontal} cuts of the formula syntax tree are
incorporated into the feature vector.

Vertical features capture top-down paths of a fixed length (e.g., three) within
the formula syntax tree. Each vertical feature, represented by a fixed-length
sequence of symbols, is assigned an index in the feature vector, with the
corresponding value indicating how many times this feature appears in the
formula.
For example, in the SMT formula \verb|(or (not (P a)) (Q a))|
we have the following paths of length 3: 
    $(\verb|or|,\verb|not|,\verb|P|)$,
    $(\verb|not|,\verb|P|,\verb|a|)$,
    $(\verb|or|,\verb|Q|,\verb|a|)$,
each one appearing once in the formula.
 
Similarly, horizontal features represent horizontal cuts in the syntax tree,
providing a simplified view of subformulas. For each subformula, a horizontal
feature consists of the subformula’s head symbol along with the top-level
symbols of its arguments.
For example, the horizontal feature corresponding to the above SMT formula
\verb|(or (not (P a)) (Q a))|, is $(\verb|or|,\verb|not|,\verb|Q|)$.
Other horizontal features corresponding to subformulas are
$(\verb|not|,\verb|P|)$, $(\verb|P|,\verb|a|)$, and $(\verb|Q|,\verb|a|)$.
Again, each horizontal feature is assigned an index in the feature vector, with
its value representing the number of occurrences in the formula.

Additionally, we incorporate the types of uninterpreted symbols into the
feature names. Uninterpreted symbols always have an arrow type of the form 
$(\tau_0\rightarrow\tau_1\rightarrow\dots\rightarrow\tau_n)$. 
Consequently, these types can be represented as sequences of symbols, analogous
to vertical and horizontal features.
Sequences representing types are simply appended to the vertical and horizontal features following each uninterpreted symbol.

The mapping of features to vector indices is performed using the
general-purpose hashing function \emph{sdbm}. The \emph{sdbm} hash function is
a widely used string hashing algorithm known for its simplicity and effective
distribution properties. It operates using bitwise shifts and arithmetic
operations to efficiently compute hash values where each character of the
string contributes to the final hash value. Given a non-empty sequence of
symbols $s=(s_0,\dots,s_{n})$, the function is defined as $h_{i+1}(s) = s_i +
(h_i(s)\ll 6) + (h_i(s)\ll 16) - h_i(s)$ where $h_0(s) = 0$ and where $\ll$
denotes the bitwise shift operation performed on 32-bit long unsigned integers.
The hash value of $s$ is $h_{n+1}(s)$. The symbols $s_i$ are typically
represented by ASCII codes, but in our case, we use cvc5's internal kind IDs.

In principle, any well-distributed hash function could be used here, as learning
performance does not depend on specific properties of \emph{sdbm}. We selected
\emph{sdbm} primarily for its simplicity and efficiency, but other common
functions such as \emph{djb2}, FNV, or MurmurHash would serve equally
well~\cite{DJBhash,FNV,MurmurHash}. Functions like \emph{djb2}, in particular,
tend to spread strings with long common prefixes more strongly than
\emph{sdbm}, which may be advantageous in some contexts. Our choice of
\emph{sdbm} reflects a trade-off between simplicity, speed, and sufficiently
good distribution for our purposes.

The \emph{sdbm} function allows us to translate a sequence of symbols into hash
values. The final hash value is reduced to an index using the modulo function,
mapping each feature to a hashing \emph{bucket}. The desired count of buckets
is determined by the user-selected parameter. We concatenate the bag-of-words
feature vector $\alpha_q$ with two separate vectors for vertical features
$\beta_q$ and horizontal features $\gamma_q$. Here, the length of $\alpha_q$
corresponds to the count of different kinds in cvc5, while $\beta_q$ and
$\gamma_q$ have a length equal to the selected bucket counts. Finally, the SMT
formula $q$ is represented by the vector $\varphi_q$, produced as the
concatenation of vectors $\alpha_q$, $\beta_q$, and $\gamma_q$.

All symbols in features are represented by their cvc5 kind IDs, abstracting
away from the different names of uninterpreted symbols. In the context of the
first-order logic UF theory, this means that all function and predicate symbols
are indistinguishable in feature vectors. However, the type information
included in path features enables us to distinguish function and predicate
symbols of different arities.

The feature vector representation of formulas is computed by our version of
cvc5 after internal formula normalization. Specifically, formulas are first
converted to negation normal form (NNF) and skolemized before being sent to
instantiation modules. The NNF transformation moves quantifiers to the
outermost level, propagates negations inward, and reduces logical connectives
to only $\land$, $\lor$, and $\lnot$. As a result, the bag-of-words feature
vectors become largely independent of the original order of quantifiers,
negations, and Boolean connectives in the formula. This normalization mitigates
the risk of large equivalence classes of syntactically different but
semantically equivalent formulas, making the simple count-based features
effective for distinguishing formulas in practice. As a potential direction for
future work, one could consider transforming formulas into a stricter normal
form, such as conjunctive normal form (CNF), before computing feature vectors.
This may further reduce sensitivity to equivalent formulas with different
Boolean encodings.

\subsection{Extraction of training examples}%
\label{sec:ml-samples}

The usefulness of a quantified formula might depend on the context of the problem being
solved.  Hence it is essential to allow our models to produce
context-dependent predictions. We achieve this by embedding the problem
features within the feature vector. First, we represent the problem being
solved ($P$) by the vector $\varphi_P$, obtained as the sum $\sum_{f\in
P}{\varphi_f}$ where $\varphi_f$ is the feature vector of each formula $f$
asserted in problem $P$. Second, the context vector $\varphi_P$ and the
quantifier vector $\varphi_q$ are concatenated into the double-length vector
$(\varphi_P,\varphi_q)$ representing the quantified formula $q$ when solving
problem $P$. This enables problem-specific predictions.
To generate training examples, we utilize cvc5's option to dump instantiations
(\texttt{dump-instantiations}) employed in solving a problem. When this option
is combined with the \texttt{produce-proofs} option, only instantiations
necessary for the proof are printed. We use this differentiation to classify
formulas as positive or negative as follows. Following a successful (unsat) run
of cvc5 on problem $P$, we gather all the quantified formulas $q$ processed by
instantiation modules during the run. To construct training examples, we label a feature
vector $(\varphi_P,\varphi_q)$ as \emph{positive} if $q$ generated an
instantiation required by the proof of $P$, otherwise we label it as \emph{negative}.
Training examples can thus be extracted from both ML-guided and unguided runs.

\subsection{Model training}%
\label{sec:ml-training}

To construct a training dataset, we collect labeled training vectors from a
large set of successful runs of cvc5. Note that no training data is extracted
from unsuccessful runs, as there is no reference point to label processed
formulas as positive or negative. It is essential to collect a substantially
large amount of training data to enhance the generalization capabilities of the
model. The gradient boosting decision tree models can be easily constructed
using the LightGBM library~\cite{LightGBM}, which is known to handle large training data well.
A model consists of a sequence of decision trees, where each tree is
trained to correct any imprecision introduced by the previous trees in the
sequence. The number of different trees in the model is one of the many
LightGBM \emph{hyperparameters} that influence the process of model training and also
the performance of the resulting model.

In order to prevent overfitting of the model to the training data, we split the
data into training and development sets. This split is done at the problem
level, ensuring that all training vectors from a single problem contribute to
the same set. We use binary classification as the model objective and we train
several models on the training set with various values of selected
hyperparameters. Out of these models, we select the model with the best
performance on the development set.

The performance of the model on the development set is measured in terms of
prediction accuracy. 
That is, all vectors from the development set are evaluated using the trained
model, and the predicted labels are compared with the expected labels from the
development set. While processing a redundant quantified formula cannot
completely prevent an SMT solver from finding a solution, omitting to process
an important formula can ultimately close the door to success. Hence, it is
essential for models for quantifier selection in SMT to favor recognition of
positive examples, that is, to minimize \emph{false negative} errors. We achieve
this by computing separately the accuracies on positive and negative
development examples, denoted $\mathit{pos}$ and $\mathit{neg}$ respectively,
and selecting as the model with the best value of
$2\cdot\textit{pos}+\textit{neg}$ as the final model. In this way, we favor
models with better performance on positive training examples.

\subsection{Model integration}%
\label{sec:ml-using}

Since LightGBM provides a C++ interface for model predictions, it can be easily
integrated into the cvc5 codebase, which is also written in C++. The interface
offers methods to load a model and compute the prediction of a quantified
formula represented by a feature vector. Since we employ binary classification
for the model objective, the prediction function returns a floating-point value
between \numprint{0} and \numprint{1}. Typically, predictions below the fixed
threshold of $0.5$ are considered negative, and those above are considered
positive. This threshold choice is standard in binary classification tasks
since it corresponds to the Bayes-optimal decision rule under balanced class
weights and symmetric error costs~\cite{murphy2012ml}. 

However, we do not directly utilize the predicted values in this manner.
Instead, each time we predict a quantified formula within an instantiation
module, we generate a random number between \numprint{0} and \numprint{1} to
serve as the threshold for comparison with the prediction. This approach allows
us to process formulas scored below $0.5$, ensuring that every formula will
eventually be processed with some non-zero probability. This implementation
further guards against potential irreparable errors resulting from false
negative predictions. Moreover, this approach sets our method apart from a
related technique known as \emph{offline premise
selection}~\cite{abs-1108-3446,GoertzelJKOPU22}, where formulas are filtered in
advance and never passed to the solver.

The two most prominent modules for quantifier selection in cvc5 are
\emph{enumerative instantiation} and \emph{e-matching}. We implement our
machine learning guidance for them, as well as for the modules relying on
\emph{conflict-based quantifier instantiation} and \emph{finite model finding}.

\section{Experimental evaluation of quantifier selection for cvc5}
\label{sec:exp}

In this section, we conduct a straightforward and reproducible experiment to
construct an ML-enhanced portfolio of strategies using the methodology
presented in Section~\ref{sec:cvc5ml}. This approach is applied to two
benchmark datasets, Mizar's MPTP and HOL4's GRUNGE, which are described in
Section~\ref{sec:exp-datasets}. Section~\ref{sec:exp-baseline} outlines the
construction of state-of-the-art cvc5 portfolios, each consisting of six
strategies, which serve as the baseline for our experiments.

The experiment consists of the following steps.
\begin{enumerate}
    \item \emph{Evaluate} baseline strategies on the training data and collect training examples (Section~\ref{sec:exp-models}).
    \item \emph{Build} models for each strategy data separately and create ML-enhanced strategies (Section~\ref{sec:exp-models}).
    \item \emph{Evaluate ML} strategies on the development set and identify the top-performing ones (Section~\ref{sec:exp-eval}).
    \item \emph{Validate} the performance of the top development strategies on the holdout set (Section~\ref{sec:exp-eval}).
\end{enumerate}
 
In Section~\ref{sec:exp-features}, we conclude by assessing the impact of
\emph{path features} from Section~\ref{sec:ml-vectors}.

All solver runs performed in this experiments are launched with the
\textbf{time limit of \numprint{60} seconds} per strategy and problem.%
\footnote{All the experiments were performed on machines with two AMD EPYC 7513
32-Core processors @ 3680 MHz and with 514 GB RAM.}
Hence, a portfolio of six strategies can take up to \numprint{360} seconds to solve each problem.
Note that this includes the time used by the trained ML predictor, that is, the
times used for the guided and unguided runs are fully comparable and there is
no hidden time involved by running the predictor remotely and/or on a GPU.

\subsection{Benchmark datasets}
\label{sec:exp-datasets}

We use two different benchmark datasets for our experiments: Mizar's MPTP~\cite{Urb03,Urb04-MPTP0,Urban06} and
HOL4's GRUNGE~\cite{BrownGKSU19}. 
Both datasets originate from translations of large mathematical
libraries into first-order logic, specifically into the UF theory in the
context of SMTs. They contain various mathematical problems that, while
distinct, share a certain level of similarity due to their common mathematical
domain. This contrasts with general-purpose benchmarks such as TPTP and
SMT-LIB, which include a diverse range of problems from different domains. The
inherent similarity within our datasets is crucial for our ML-based approach,
as it facilitates knowledge transfer and improves the ability to generalize
across related problems.

In both cases, we split the dataset into \emph{training}, \emph{development} (or just \emph{devel}),
and \emph{holdout} parts. The training set is used to extract training
examples, while the development set is used to tune model hyperparameters and
select the best strategies. Only the final portfolio of strategies that perform
best on the development set is then evaluated on the holdout set.

\paragraph{Mizar MML dataset}

The Mizar Mathematical Library (MML)~\cite{BancerekBGKMNP18} stands as one of
the earliest extensive repositories of formal mathematics, encompassing a broad
spectrum of lemmas and theorems across various mathematical domains. We utilize
translations of MML problems into first-order logic, a process facilitated by
the MPTP system~\cite{Urb03,Urb04-MPTP0,Urban06}. The MPTP benchmark has emerged as a
valuable resource for machine learning research, offering a diverse collection of
related problems~%
\cite{DBLP:conf/frocos/RawsonR19,JakubuvU19,DBLP:conf/frocos/Suda21,DBLP:conf/tableaux/RawsonR21,DBLP:journals/corr/abs-2303-15642,ChvalovskyKPU23}.
In our work, we focus on the easier \emph{bushy} variants of MPTP problems, wherein
premises are partially filtered externally beforehand.

We use a train/development/holdout split from other experiments in the
literature~\cite{JakubuvCGKOP00U23,GoertzelCJOU21} to allow a competitive
evaluation.
This splits the whole set of Mizar problems into the \emph{training},
\emph{development}, and \emph{holdout} subsets, using a 90~:~5~:~5 ratio,
yielding \numprint{52125} problems in the training set,
\numprint{2896} in devel, and \numprint{2896} problems in the holdout set.
We use the \emph{training} set to collect training examples for machine learning, 
and we use the \emph{development} set to select the best model out of several
candidates trained with different hyperparameters.
The best portfolio constructed on the development is henceforth evaluated
on the \emph{holdout} set and compared with the baseline portfolio.

\paragraph{The GRUNGE dataset}
To see how general our methods are, we evaluate them also on the \emph{GRUNGE} (Grand Unified ATP Challenge) dataset~\cite{BrownGKSU19}.
GRUNGE is derived from the HOL4~\cite{SlindN08,GordonMelham93} standard library
(Kananaskis-12 version), consisting of \numprint{15733} formulae,
including \numprint{8} axioms, \numprint{2294} definitions, and
\numprint{13431} theorems. The dataset contains \numprint{12140}
theorem-proving problems, structured into first-order and higher-order
logic (HOL) variants: %
untyped first-order (FOF),
monomorphic first-order (TF0), %
polymorphic first-order (TF1), %
monomorphic higher-order
(TH0), and %
polymorphic higher-order (TH1). The dataset
targets ATP evaluation across these different logical formalisms 
using appropriate TPTP formats.

GRUNGE uses two translation families: first-order encodings (FOF-I,
TF0-I, TF1-I, TH0-I, TH1-I) and set-theoretic encodings (FOF-II, TF0-II, TH0-II).
The dataset also contains two types of problem versions: \emph{bushy} and
\emph{chainy}. The bushy version includes only the minimal set of
dependencies required for the HOL proof, whereas the chainy version
includes all prior theorems in the library order. The chainy problems
are often much larger. They emulate real-world theorem-proving scenarios where
ATPs must select relevant premises from a large set of possible
dependencies. The difference in performance between these versions
provides insight into the effectiveness of premise selection
strategies in large-theory settings.

An empirical evaluation across \numprint{19} ATP systems in~\cite{BrownGKSU19} showed
that \numprint{7412} out of \numprint{12094} %
bushy problems (\numprint{61.1}\%)
could be solved, with Leo-III solving \numprint{7090}, CVC4 \numprint{5709},
E 2.2 \numprint{5118}, and Vampire 4.3 \numprint{5929}. TacticToe, an
internal HOL4 prover, solved \numprint{5327} out of \numprint{8855}
chainy problems (\numprint{60.2}\%). Combined, ATPs and TacticToe solved
\numprint{8840} problems (\numprint{72.8}\%). The first-order encodings
generally led to higher success rates, but some provers (e.g.,
Vampire 4.3, SPASS 3.9) performed better with set-theoretic
translations.

GRUNGE was also used in CASC-27 competition~\cite{casc27} within its Large Theory
  Batch (LTB) division~\cite{Sutcliffe09}, supporting FOF, TF0, TF1, TH0, and
  TH1 logic variants. It provides a unified benchmark for ATPs, enabling direct
performance comparisons across different logical systems. The dataset
structure, together with problem translations and theorem
dependencies, enables comprehensive testing of automated reasoning
techniques in large-theory settings.

The GRUNGE dataset consists of \numprint{12094} problems. %
Again, we use a training/development/holdout split to assess the overfitting of
our models. Since the GRUNGE dataset is smaller than Mizar, we use a ratio of
80~:~10~:~10 to obtain a reasonably sized holdout set. This gives us
\numprint{9674} training problems, and \numprint{1210} problems in
both the development and holdout sets. With the smaller dataset, we also test
the ability of our method to learn from a reduced amount of training data.

\subsection{Baseline strategies and portfolios}%
\label{sec:exp-baseline}

\begin{figure*}[t]
\footnotesize
\begin{center}
\def\arraystretch{1.3}%
\setlength\tabcolsep{1em}%
\begin{tabular}{l|l}
   \emph{strategy} & \emph{cvc5 command line options}
\\\hline\hline
\casc{ 1} & \texttt{\textbf{full-saturate-quant}}       \quad\texttt{decision=internal}         \quad\texttt{simplification=none}       \quad \texttt{no-inst-no-entail} \quad \texttt{no-cbqi} \\\hline
\casc{ 2} & \texttt{\textbf{full-saturate-quant}}       \quad\texttt{\textbf{no-e-matching}}                                                                                               \\\hline
\casc{ 3} & \texttt{\textbf{full-saturate-quant}}       \quad\texttt{\textbf{no-e-matching}}             \quad \texttt{enum-inst-sum}                                                             \\\hline
\casc{ 4} & \texttt{\textbf{finite-model-find}  }       \quad\texttt{uf-ss=no-minimal}                                                                                            \\\hline
\casc{ 5} & \texttt{\textbf{full-saturate-quant}}       \quad\texttt{\textit{multi-trigger-when-single}}                                                                                                                            \\\hline
\casc{ 6} & \texttt{\textbf{full-saturate-quant}}       \quad\texttt{\textit{trigger-sel=max}}                                                                                                                                      \\\hline
\casc{ 7} & \texttt{\textbf{full-saturate-quant}}       \quad\texttt{\textit{multi-trigger-when-single}} \quad \texttt{\textit{multi-trigger-priority}}                                                                                      \\\hline
\casc{ 8} & \texttt{\textbf{full-saturate-quant}}       \quad\texttt{\textit{multi-trigger-cache}}                                                                                                                                  \\\hline
\casc{ 9} & \texttt{\textbf{full-saturate-quant}}       \quad\texttt{prenex-quant=none}                                                                                                                                    \\\hline
\casc{10} & \texttt{\textbf{full-saturate-quant}}       \quad\texttt{enum-inst-interleave}      \quad \texttt{decision=internal}                                                                                           \\\hline
\casc{11} & \texttt{\textbf{full-saturate-quant}}       \quad\texttt{\textit{relevant-triggers}}                                                                                                                                    \\\hline
\casc{12} & \texttt{\textbf{finite-model-find}  }                                               \quad \texttt{sort-inference}         \quad \texttt{uf-ss-fair}               \\\hline
\casc{13} & \texttt{\textbf{full-saturate-quant}}       \quad\texttt{pre-skolem-quant=on}                                                                                                                                  \\\hline
\casc{14} & \texttt{\textbf{full-saturate-quant}}       \quad\texttt{\textbf{cbqi-vo-exp}}                                                                                                                                          \\\hline
\casc{15} & \texttt{\textbf{full-saturate-quant}}       \quad\texttt{no-cbqi}                                                                                                                                              \\\hline
\casc{16} & \texttt{\textbf{full-saturate-quant}}       \quad\texttt{macros-quant}              \quad \texttt{macros-quant-mode=all}                                                 
\\\hline\hline
\grk{1} & 
\texttt{\textbf{full-saturate-quant}} \quad
\texttt{\textbf{cbqi-vo-exp}}  \quad
\texttt{\textit{\textbf{relational-triggers}}} \quad
\texttt{cond-var-split-quant=agg} 
\\\hline
\grk{2} & 
\texttt{\textbf{full-saturate-quant}} \quad
\textbf{\texttt{cbqi-vo-exp}} \quad
\texttt{\textit{\textbf{relevant-triggers}}} \quad
\texttt{\textit{\textbf{multi-trigger-priority}}}
\\ & 
\texttt{ieval=off} \quad
\texttt{no-static-learning} \quad
\texttt{miniscope-quant=off}
\\\hline
\grk{3} & 
\texttt{\textbf{full-saturate-quant}} \quad
\texttt{\textit{\textbf{multi-trigger-priority}}} \quad
\texttt{\textit{\textbf{multi-trigger-when-single}}}
\\ & 
\texttt{term-db-mode=relevant} 

\end{tabular}
\end{center}
\normalsize
\caption{Selected cvc5's strategies from the CASC portfolio.}
\label{fig:strats-casc}
\end{figure*}

We employ a state-of-the-art portfolio of cvc5 strategies as a baseline to evaluate
our machine learning approach. 
We incorporate all 16 strategies available in
the CASC competition portfolio of cvc5.%
\footnote{\url{https://github.com/cvc5/cvc5/blob/cvc5-1.1.1/contrib/competitions/casc/run-script-cascj11-fof}}
These strategies represent the state-of-the-art methods provided by cvc5 to
tackle first-order problems.
A portfolio consists of strategies that
are typically executed sequentially, each for a fraction of the overall time
limit. The first successful strategy returns the solution and terminates the
portfolio execution. The CASC competition focuses on problems in automated
theorem proving, and cvc5 has recently demonstrated outstanding performance on
problems from theorem proving systems~\cite{GoertzelJKOPU22}. We utilize the
FOF (first-order formulas) portfolio, specifically designed for problems in the
theory of uninterpreted functions (UF), which aligns well with our
intended application on Mizar and GRUNGE problems expressed in the same logic.

In addition to the \numprint{16} CASC strategies, we also incorporate
\numprint{3} publicly available
strategies from our previous work~\cite{cicm24}.
These strategies%
\footnote{\url{https://github.com/ai4reason/cvc5_grackle_mizar}}
were developed recently for the Mizar MML problems using the Grackle strategy
invention system~\cite{cicm24,HulaJJK22}.
The Grackle strategies in that work were tailored for slightly different versions of Mizar
problems, employing a more precise method of premise selection. However, they
also perform well on the \emph{bushy} version of Mizar problems used in our work here.

The strategies, denoted \casc{n} and \grk{n}, are detailed in
Figure~\ref{fig:strats-casc}.
In both cases, strategies are
described in terms of their cvc5 command line options, either as a Boolean \texttt{flag}
or as an \texttt{option=value} pair. Common options appearing in more than one
strategy are typeset in \texttt{\textbf{bold}}. Options adjusting the trigger
behavior of the e-matching module are highlighted in
\texttt{\textbf{\textit{bold-italics}}}. Note that almost all the strategies
activate enumerative instantiation using \texttt{full-saturate-quant}, while
keeping e-matching and conflict-based quantifier instantiation (\texttt{cbqi})
turned on by default. The option \texttt{cbqi-vo-exp}, which activates an
experimental mode for variable ordering, often significantly boosts strategy
performance.
It is worth noting that \grk{1} and \grk{2} are
proper extensions of \casc{14}, while \grk{3} is a proper extension of
\casc{7}. The core of a successful cvc5 strategy for Mizar problems appears to
be enumerative instantiation with appropriate adjustments of triggers for
e-matching.

One common way to assess the performance of a strategy portfolio is via the
\emph{greedy cover} sequence. This sequence is generated by first selecting the
best-performing individual strategy and marking the problems it solves.
Subsequently, the next best strategy—based on the remaining unsolved
problems—is chosen. The process continues iteratively until no further
strategies can improve coverage. Through this approach, the greedy cover
identifies strategies that complement each other, thereby maximizing the total
number of solved problems. This method approximates the NP-complete
\emph{set cover} problem~\cite{GareyJ79}, which in principle could be solved
exactly but in practice rarely yields significant improvements
~\cite{febbd0fd-43e1-3b6c-bf72-2de916079418,DBLP:journals/corr/abs-2403-12869}.
Hence, the greedy cover construction approximates the virtual best portfolio
closely enough for our purposes.

We evaluate all \numprint{19} baseline strategies on the \emph{development} set
to establish the state-of-the-art baseline portfolio for our datasets. We
select the best portfolio of size \numprint{6} separately for each dataset,
which are henceforth considered \emph{baseline portfolios}. The performance of
the baseline portfolios on the holdout dataset split is presented in
Table~\ref{tab:baseline}, with Mizar on the left and GRUNGE on the right.

\begin{table}[h!]
\caption{Baseline portfolios performance without ML.}\label{tab:baseline}
\begin{minipage}{0.49\columnwidth}
\begin{center}
\def\arraystretch{1.1}%
\setlength\tabcolsep{0.7em}%
\begin{tabular}{l||rrl|l}
   \emph{strategy} & \emph{solves} & $+$\emph{new} & \emph{adds} & $=$ \emph{total} 
\\\hline
\sparrow     & $\numprint{1472}  $ & $+\numprint{1472}  $ & $-                  $ & $=\numprint{1472}  $ \\
\chickadee   & $\numprint{1400}  $ & $+\numprint{77}    $ & $+\numprint{5.23}\% $ & $=\numprint{1549}  $ \\
\casc{13}    & $\numprint{1434}  $ & $+\numprint{45}    $ & $+\numprint{2.91}\% $ & $=\numprint{1594}  $ \\
\buzzard     & $\numprint{1433}  $ & $+\numprint{22}    $ & $+\numprint{1.38}\% $ & $=\numprint{1616}  $ \\
\casc{ 4}    & $\numprint{782}   $ & $+\numprint{18}    $ & $+\numprint{1.11}\% $ & $=\numprint{1634}  $ \\ \cline{5-5}
\casc{16}    & $\numprint{1381}  $ & $+\numprint{9}     $ & $+\numprint{0.55}\% $ & \multicolumn{1}{||l||}{$\mathbf{=\numprint{1643}}$} \\ \cline{5-5}
\end{tabular}

\vspace{2mm}
\small
\setlength\tabcolsep{0.5em}%
\begin{tabular}{|ll|} 
    \hline
    \textbf{benchmark:}     & Mizar/holdout with \numprint{2896} problems \\
    \textbf{strategies:}    & 6 baseline strategies best on Mizar/devel\\
    \hline
\end{tabular}
\normalsize
\end{center}
\end{minipage}
\begin{minipage}{0.49\columnwidth}
\begin{center}
\def\arraystretch{1.1}%
\setlength\tabcolsep{0.7em}%
\begin{tabular}{l||rrl|l}
   \emph{strategy} & \emph{solves} & $+$\emph{new} & \emph{adds} & $=$ \emph{total} 
\\\hline
\casc{ 1}     & $\numprint{489}   $ & $+\numprint{489}   $ & $-       $ & $=\numprint{489}   $ \\
\casc{ 6}     & $\numprint{465}   $ & $+\numprint{40}    $ & $+\numprint{8.18}\% $ & $=\numprint{529}   $ \\
\casc{ 5}     & $\numprint{478}   $ & $+\numprint{13}    $ & $+\numprint{2.46}\% $ & $=\numprint{542}   $ \\
\casc{13}     & $\numprint{479}   $ & $+\numprint{5}     $ & $+\numprint{0.92}\% $ & $=\numprint{547}   $ \\
\buzzard      & $\numprint{485}   $ & $+\numprint{3}     $ & $+\numprint{0.55}\% $ & $=\numprint{550}   $ \\ \cline{5-5}
\casc{ 4}     & $\numprint{199}   $ & $+\numprint{2}     $ & $+\numprint{0.36}\% $ & \multicolumn{1}{||l||}{$\mathbf{=\numprint{552}   }$} \\ \cline{5-5}
\end{tabular}

\vspace{2mm}
\small
\setlength\tabcolsep{0.5em}%
\begin{tabular}{|ll|} 
    \hline
    \textbf{benchmark:}     & GRUNGE/holdout with \numprint{1210} problems \\
    \textbf{strategies:}    & 6 baseline strategies best on GRUNGE/devel\\
    \hline
\end{tabular}
\footnotesize
\end{center}
\end{minipage}
\normalsize
\end{table}

The column \emph{strategy} lists the strategy names in the order determined by
the greedy cover. The column \emph{solves} shows the number of problems solved
individually by each strategy. The remaining columns indicate the contribution
of each strategy to the overall portfolio performance. The column \emph{+new}
shows how many new problems each strategy adds to the portfolio. The column
\emph{total} denotes the cumulative number of problems solved by the portfolio
up to that point, where \emph{total} equals \emph{new} plus the \emph{total}
from the previous row. Finally, the column \emph{adds} presents the value of
\emph{new} as a percentage of the current portfolio performance.  For instance,
    \grk{3} adds \numprint{77} problems to the currently solved \numprint{1472},
representing $5.23\%$. The overall performance of the portfolio is reflected
in the bottom highlighted value of the \emph{=total} column. The description
below the table provides basic portfolio information to facilitate orientation
in the tables.

We can observe that our choice of a portfolio size of \numprint{6} is
reasonable, as adding more strategies would not significantly improve
performance. On Mizar, the remaining strategies contribute only \numprint{41} additional
problems, while on GRUNGE, they add just \numprint{3}. Notably, the best strategies
differ between the datasets, with only \casc{13}, \casc{4}, and \grk{2}
appearing in both portfolios. While the Grackle strategies \grk{n} dominate the
Mizar baseline portfolio, only \grk{2} is included in the GRUNGE baseline. On
Mizar, the Grackle strategies enhance the performance of CASC strategies by
$\SI{2.46}\percent$, whereas on GRUNGE, all problems solved by Grackle
strategies are already covered by other CASC strategies. This reflects the fact
that the Grackle strategies were specifically designed for Mizar. A similar
attempt to invent specialized strategies for GRUNGE using Grackle was made but
did not yield significant results yet. The differences in baseline portfolio
structures also suggest a certain level of dissimilarity between the datasets.

Upon inspecting the strategy list in Figure~\ref{fig:strats-casc}, we can
conclude that the core of a successful cvc5 strategy for both Mizar and GRUNGE
problems appears to be enumerative instantiations with appropriate adjustments
to triggers for e-matching. It is worth noting that \casc{4} is the only
strategy that does not rely on enumerative instantiations or e-matching,
instead using finite model finding~\cite{reynolds-cade13}. While finite model
finding is not expected to perform optimally on unsatisfiable problems, it can
still be useful for solving specific instances.

\subsection{Training data and model building}
\label{sec:exp-models}

Training data is extracted easily from successful solver runs as our extension
of cvc5 directly outputs labeled training vectors.
We evaluate the baseline portfolio strategies on the training data and extract
training examples from each strategy separately. This allows us to train a
distinct classification model for each strategy. For consistency, we use the
strategy's name to denote its corresponding model. For example, the model
trained on the training examples generated by \grk{2} is also referred to as
\grk{2}.

To assess the impact of \emph{path features}, we train three models for each
strategy, varying the number of \emph{buckets} in the feature vectors: \numprint{0}, and
\numprint{512}, and \numprint{1024} buckets. Specifically, the value of \numprint{0} corresponds to omitting
path features entirely, aligning with the approach used in our ECAI'24
paper~\cite{JakubuvJPU24}. The values \numprint{512} and \numprint{1024} were chosen based on
minimal preliminary experiments.
For the vertical path features, we always choose the path of length \numprint{3}.

To influence the quality of the trained model, several hyperparameters can be
adjusted during training. We always train a LightGBM model with \numprint{100} trees and
unlimited tree depth while varying the number of leaves, as this is the most
impactful hyperparameter in our case.
Specifically, we experiment with leaf counts ranging from \numprint{64} to \numprint{2048},
considering all powers of \numprint{2} and intermediate values ($2^N$ and $2^{N+0.5}$).
We then evaluate the accuracy of the trained models on the development data and
select the one that performs best.
This accuracy evaluation is performed on the development vectors, which remain
the same regardless of the number of leaves. This eliminates the need to launch
cvc5 for each model.
This process is performed for each combination of baseline strategy and bucket
count, yielding a single model for each strategy and bucket pair.

\subsection{Evaluation of ML-enhanced portfolios}
\label{sec:exp-eval}

As a result of the model training, we obtain 18 different models (\numprint{6}
strategies $\times$ \numprint{3} bucket sizes) per benchmark dataset, each
identified by its source baseline strategy and bucket count. In this section,
we always pair a baseline strategy $S$ with the model trained on training data
generated by $S$. Combinations of strategies with models trained on data from
different strategies are explored in Section~\ref{sec:cross}. Thus, each model
uniquely determines the strategy it can be combined with, resulting in 18
distinct ML-enhanced strategies. We evaluate all of them on the development set
and construct a final greedy cover portfolio of size \numprint{6}. This
portfolio closely approximates the virtual best portfolio of size \numprint{6}
on the development set. In what follows, we evaluate this final portfolio on
the holdout set.

We present the results for Mizar and GRUNGE separately, beginning with Mizar in
Table~\ref{tab:ml:mizar}. The table on the \emph{left} shows the performance of
the final portfolio on the holdout set. The table on the \emph{right} presents
the virtual best portfolio on the holdout set. It is obtained by evaluating all
available strategies and constructing the greedy cover on the holdout set. Comparing the
two indicates the degree of overfitting in our portfolio construction: a small
gap shows that the portfolio closely approximates the best performance,
whereas a large gap would suggest room for improvement.

\begin{table}[h!]
\caption{Results on Mizar/holdout with ML-enhanced strategies.}\label{tab:ml:mizar}

\begin{minipage}{0.49\columnwidth}
\begin{center}
\def\arraystretch{1.1}%
\setlength\tabcolsep{0.3em}%
\begin{tabular}{ll||lll|l}
    \emph{strategy} & \emph{buckets} & \emph{solves} & $+$\emph{new} & \emph{adds} & $=$ \emph{total} 
\\\hline
\sparrow     &  \numprint{1024}    & $\numprint{1845}  $ & $+\numprint{1845}  $ & $-       $ & $=\numprint{1845}  $ \\
\chickadee   &  \numprint{512}     & $\numprint{1771}  $ & $+\numprint{86}    $ & $+\numprint{4.66}\% $ & $=\numprint{1931}  $ \\
\buzzard     &  \numprint{512}     & $\numprint{1755}  $ & $+\numprint{44}    $ & $+\numprint{2.28}\% $ & $=\numprint{1975}  $ \\
\buzzard     &  \numprint{0}       & $\numprint{1719}  $ & $+\numprint{15}    $ & $+\numprint{0.76}\% $ & $=\numprint{1990}  $ \\
\casc{13}    &  \numprint{1024}    & $\numprint{1692}  $ & $+\numprint{12}    $ & $+\numprint{0.60}\% $ & $=\numprint{2002}  $ \\\cline{6-6}
\chickadee   &  \numprint{1024}    & $\numprint{1751}  $ & $+\numprint{10}    $ & $+\numprint{0.50}\% $ & \multicolumn{1}{||l||}{$\mathbf{=\numprint{2012}  }$} \\\cline{6-6}
\end{tabular}

\vspace{2mm}
\small
\setlength\tabcolsep{0.2em}%
\begin{tabular}{|ll|} 
\hline
\textbf{benchmark:}     & Mizar/holdout with \numprint{2896} problems \\
\textbf{models:}        & model $X$ is trained on data from strategy $X$ \\
\textbf{strategies:}    & 6 best selected on Mizar/devel\\
\textbf{improvement:}   & \textbf{$+\SI{22.46}\percent$} over the baseline in Table~\ref{tab:baseline} (left) \\
\hline
\end{tabular}
\normalsize
\end{center}
\end{minipage}
\quad
\begin{minipage}{0.49\columnwidth}
\begin{center}
\def\arraystretch{1.1}%
\setlength\tabcolsep{0.3em}%
\begin{tabular}{ll||lll|l}
\emph{strategy} & \emph{buckets} & \emph{solves} & $+$\emph{new} & \emph{adds} & $=$ \emph{total} 
\\\hline
\sparrow    & \numprint{1024}       & $\numprint{1845}  $ & $+\numprint{1845}  $ & $-       $ & $=\numprint{1845}  $ \\
\chickadee  & \numprint{0}          & $\numprint{1733}  $ & $+\numprint{86}    $ & $+\numprint{4.66}\% $ & $=\numprint{1931}  $ \\
\buzzard    & \numprint{512}        & $\numprint{1755}  $ & $+\numprint{41}    $ & $+\numprint{2.12}\% $ & $=\numprint{1972}  $ \\
\chickadee  & \numprint{512}        & $\numprint{1771}  $ & $+\numprint{24}    $ & $+\numprint{1.22}\% $ & $=\numprint{1996}  $ \\
\sparrow    & \numprint{0}          & $\numprint{1809}  $ & $+\numprint{17}    $ & $+\numprint{0.85}\% $ & $=\numprint{2013}  $ \\\cline{6-6}
\buzzard    & \numprint{1024}       & $\numprint{1751}  $ & $+\numprint{13}    $ & $+\numprint{0.65}\% $ & \multicolumn{1}{||l||}{$\mathbf{=\numprint{2026}  }$} \\\cline{6-6}
\end{tabular}

\vspace{2mm}
\small
\setlength\tabcolsep{0.2em}%
\begin{tabular}{|ll|} 
\hline
\textbf{benchmark:}     & Mizar/holdout with \numprint{2896} problems \\
\textbf{models:}        & model $X$ is trained on data from strategy $X$ \\
\textbf{strategies:}    & \text{virtual best portfolio on Mizar/holdout} \\
\textbf{gap:}   & \textbf{$+\SI{0.7}\percent$} over the development portfolio (left) \\
\hline
\end{tabular}
\end{center}
\end{minipage}
\normalsize
\end{table}

First, we observe a \textbf{significant improvement of $+\SI{22.46}\percent$ over the
Mizar baseline portfolio} from Table~\ref{tab:baseline}, increasing the number
of solved problems from \numprint{1643} to \numprint{2012}. This is a strong
result, as the baseline portfolio represents the state-of-the-art performance
in SMTs.
Next, strategies incorporating path features dominate the portfolio, with only
\grk{2} appearing without them. Interestingly, \grk{2} and \grk{3} are included twice with
different bucket sizes, completely replacing \casc{4} and \casc{16} from the baseline
portfolio. This suggests that different bucket sizes offer a degree of
complementarity.
Moreover, overfitting to the development set is minimal, as the optimal
portfolio improves the result by only $+\SI{0.7}\percent$. The optimal
portfolio consists exclusively of Grackle strategies, each appearing twice with
different bucket sizes. 
Notably, none of the baseline strategies without ML, which were also considered
during the construction of both portfolios, are included among the top six
strategies. Furthermore, they contribute only a minimal improvement, as we will
discuss in Section~\ref{sec:exp-features}.

We proceed with results on GRUNGE in Table~\ref{tab:ml:grunge}.

\begin{table}[h!]
\caption{Results on GRUNGE/holdout with ML-enhanced strategies.}\label{tab:ml:grunge}

\begin{minipage}{0.49\columnwidth}
\begin{center}
\def\arraystretch{1.1}%
\setlength\tabcolsep{0.4em}%
\begin{tabular}{ll||lll|l}
\emph{strategy} & \emph{buckets} & \emph{solves} & $+$\emph{new} & \emph{adds} & $=$ \emph{total} 
\\\hline
\casc{ 1}    & \numprint{512}     & $\numprint{568}   $ & $+\numprint{568}   $ & $-       $ & $=\numprint{568}   $ \\
\casc{13}    & \numprint{1024}    & $\numprint{559}   $ & $+\numprint{30}    $ & $+\numprint{5.28}\% $ & $=\numprint{598}   $ \\
\casc{ 1}    & \numprint{1024}    & $\numprint{567}   $ & $+\numprint{9}     $ & $+\numprint{1.51}\% $ & $=\numprint{607}   $ \\
\casc{ 5}    & \numprint{1024}    & $\numprint{541}   $ & $+\numprint{8}     $ & $+\numprint{1.32}\% $ & $=\numprint{615}   $ \\
\casc{ 6}    & \numprint{0}       & $\numprint{499}   $ & $+\numprint{6}     $ & $+\numprint{0.98}\% $ & $=\numprint{621}   $ \\\cline{6-6}
\buzzard     & \numprint{512}     & $\numprint{550}   $ & $+\numprint{1}     $ & $+\numprint{0.16}\% $ & \multicolumn{1}{||l||}{$\mathbf{=\numprint{622}   }$} \\\cline{6-6}
\end{tabular}

\vspace{2mm}
\small
\setlength\tabcolsep{0.2em}%
\begin{tabular}{|ll|} 
\hline
\textbf{benchmark:}     & GRUNGE/holdout with \numprint{1210} problems \\
\textbf{models:}        & model $X$ is trained on data from strategy $X$ \\
\textbf{strategies:}    & 6 best selected on GRUNGE/devel\\
\textbf{improvement:}   & \textbf{$+\SI{12.86}\percent$} over the baseline in Table~\ref{tab:baseline} (right) \\
\hline
\end{tabular}
\normalsize
\end{center}
\end{minipage}
\begin{minipage}{0.49\columnwidth}
\begin{center}
\def\arraystretch{1.1}%
\setlength\tabcolsep{0.4em}%
\begin{tabular}{ll||lll|l}
\emph{strategy} & \emph{buckets} & \emph{solves} & $+$\emph{new} & \emph{adds} & $=$ \emph{total} 
\\\hline
\casc{1} &    \numprint{512}     & $\numprint{568}   $ & $+\numprint{568}   $ & $-       $ & $=\numprint{568}   $ \\
\casc{13} &   \numprint{1024}    & $\numprint{559}   $ & $+\numprint{30}    $ & $+\numprint{5.28}\% $ & $=\numprint{598}   $ \\
\casc{1} &    \numprint{1024}    & $\numprint{567}   $ & $+\numprint{9}     $ & $+\numprint{1.51}\% $ & $=\numprint{607}   $ \\
\casc{6} &    \numprint{512}     & $\numprint{511}   $ & $+\numprint{9}     $ & $+\numprint{1.48}\% $ & $=\numprint{616}   $ \\
\casc{5} &    \numprint{1024}    & $\numprint{541}   $ & $+\numprint{8}     $ & $+\numprint{1.30}\% $ & $=\numprint{624}   $ \\\cline{6-6}
\casc{1} &    \numprint{0}       & $\numprint{550}   $ & $+\numprint{5}     $ & $+\numprint{0.80}\% $ & \multicolumn{1}{||l||}{$\mathbf{=\numprint{629}   }$} \\\cline{6-6}
\end{tabular}

\vspace{2mm}
\small
\setlength\tabcolsep{0.2em}%
\begin{tabular}{|ll|} 
    \hline
    \textbf{benchmark:}     & GRUNGE/holdout with \numprint{1210} problems \\
    \textbf{models:}        & model $X$ is trained on data from strategy $X$ \\
    \textbf{strategies:}    & \text{virtual best portfolio on GRUNGE/holdout} \\
    \textbf{gap:}   & \textbf{$+\SI{1.13}\percent$} over the development portfolio (left) \\
    \hline
\end{tabular}
\end{center}
\end{minipage}
\normalsize
\end{table}

Again, we observe a \textbf{significant improvement of $\SI{12.86}\percent$ over the
baseline portfolio} in Table~\ref{tab:baseline}, increasing the number of solved
problems from \numprint{552} to \numprint{622}. Similarly, overfitting to the
development set is minimal, at $\SI{1.13}\percent$. The slightly less
favorable results compared to Mizar can most likely be attributed to the
smaller size of the GRUNGE benchmark, and therefore to the reduced amount of
training data.
The use of path features follows a similar pattern, with only \casc{6}
appearing without them. Notably, none of the Grackle strategies are present in
the virtual best portfolio. Interestingly, \casc{1}, the strongest strategy on
GRUNGE, appears three times in the virtual best portfolio, with all
possible bucket sizes.

\begin{figure*}[t]
   \begin{center}
      \includegraphics[width=1.0\columnwidth]{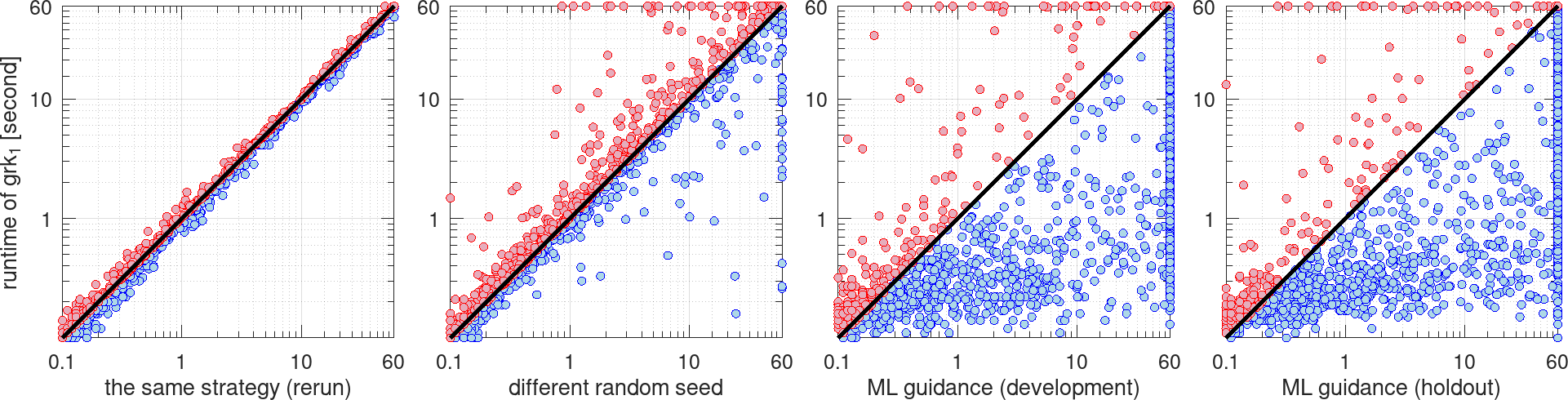}
   \end{center}
\caption{Impact of machine guidance on runtime.}%
\label{fig:runtimes}
\end{figure*}

The impact of ML-enhanced \grk{1} on runtimes on Mizar is illustrated in
Figure~\ref{fig:runtimes} through scatter plots. Each plot compares the
runtimes of two strategies, $S_1$ and $S_2$, by plotting a dot for each problem
$p$ at coordinates $(r_1, r_2)$, representing the runtimes of $S_1$ and $S_2$
on $p$. Points clustered around the diagonal line indicate similar results,
while points below the diagonal (depicted in blue) represent an improvement of
$S_2$ over $S_1$. In our context, the first strategy, $S_1$, is consistently
\grk{1} (without ML), which is compared with different strategies in various
plots.

In Figure~\ref{fig:runtimes}, the first two plots compare \grk{1} with (1)
another run of itself and (2) a run of \grk{1} with a different random seed.
The first plot illustrates the deterministic nature of the solver, showing
similar performance when launched again on the same problem. However, certain
aspects of cvc5 involve randomization, and running with a different initial
random seed can yield different results, as demonstrated by the second plot. We
observe a typical variation of around $\SI{2}\percent$ between two runs with
different random seeds.

The last two plots compare \grk{1} without ML with \grk{1} with ML on the
development set (third plot from the left) and the holdout set (rightmost
plot). Remarkably, there is no substantial difference in performance between
the development and holdout sets. Furthermore, it is evident that the impact of
ML is much more significant than that of a different random seed. Notably,
there is a large number of points below the diagonal and a significant cluster
of points at the right axis, showing problems solved due to ML.

We conclude this experiment evaluation with the observation that, in both the
Mizar and GRUNGE cases, \textbf{the best ML-enhanced strategy alone can solve more
problems in \numprint{60} seconds than the six baseline strategies together in
\numprint{360} seconds}. This represents a significant improvement achieved by
our ML strategies.

\subsection{Contribution of path features}
\label{sec:exp-features}

We conclude this section by evaluating the contribution of path features. For
each strategy, we have one model without path features (0 buckets) and two
models with path features hashed into different bucket counts. For every
possible bucket count (0, 512, 1024), we have six ML-enhanced strategies,
allowing us to directly evaluate on the holdout set. There is no need for
selection on the development set in this case because there is only one
possibility. As a result, we obtain one portfolio for each bucket count, and we
can use their performance to measure the impact of path features. The
performance of the path feature portfolios is presented in
Table~\ref{tab:ml:buckets}, for Mizar (left) and GRUNGE (right).

\begin{table}[h]
\caption{Evaluation of the impact of path features.}\label{tab:ml:buckets}

\begin{minipage}{0.49\columnwidth}
\begin{center}
\def\arraystretch{1.1}%
\setlength\tabcolsep{0.6em}%
\begin{tabular}{r||rl|rl}
    & \multicolumn{1}{c}{\emph{total}} & \multicolumn{1}{c|}{\emph{baseline}} & \multicolumn{2}{c}{\emph{solutions}} 
\\
\emph{buckets} & \emph{solved} & \emph{improvement} & \emph{gained} & \emph{lost} 
\\\hline
\numprint{0}    & $\numprint{1970}$  & $+\SI{19.9}\percent$  & $+\numprint{343}$ & $-\numprint{17}$ \\
\numprint{512}  & $\numprint{2005}$  & $+\SI{22.03}\percent$ & $+\numprint{370}$ & $-\numprint{8}$ \\
\numprint{1024} & $\numprint{2001}$  & $+\SI{21.79}\percent$ & $+\numprint{366}$ & $-\numprint{8}$ \\
\end{tabular}

\vspace{2mm}
\small
\setlength\tabcolsep{0.5em}%
\begin{tabular}{|ll|} 
    \hline
    \textbf{benchmark:}     & Mizar/holdout with \numprint{2896} problems \\
    \textbf{portfolio:}    & \numprint{6} strategies with the same bucket count\\
    \hline
\end{tabular}
\normalsize
\end{center}
\end{minipage}
\quad
\begin{minipage}{0.49\columnwidth}
\begin{center}
\def\arraystretch{1.1}%
\setlength\tabcolsep{0.6em}%
\begin{tabular}{r||rl|rl}
    & \multicolumn{1}{c}{\emph{total}} & \multicolumn{1}{c|}{\emph{baseline}} & \multicolumn{2}{c}{\emph{solutions}} 
\\
\emph{buckets} & \emph{solved} & \emph{improvement} & \emph{gained} & \emph{lost} 
\\\hline
\numprint{0}    & $\numprint{612}$  & $+\SI{10.87}\percent$ & $+\numprint{65}$ & $-\numprint{5}$ \\
\numprint{512}  & $\numprint{620}$  & $+\SI{12.32}\percent$ & $+\numprint{69}$ & $-\numprint{1}$ \\
\numprint{1024} & $\numprint{624}$  & $+\SI{13.04}\percent$ & $+\numprint{72}$ & $-\numprint{0}$ \\
\end{tabular}

\vspace{2mm}
\small
\setlength\tabcolsep{0.5em}%
\begin{tabular}{|ll|} 
    \hline
    \textbf{benchmark:}     & GRUNGE/holdout with \numprint{1210} problems \\
    \textbf{portfolio:}    & \numprint{6} strategies with the same bucket count\\
    \hline
\end{tabular}
\normalsize
\end{center}
\end{minipage}
\normalsize
\end{table}

The column \emph{total solved} presents the total number of problems solved by
each portfolio, while the column \emph{baseline improvement} shows the
improvement over the baseline in Table~\ref{tab:baseline}, expressed as a
percentage. The \emph{solutions} columns represent the difference in the number
of solved problems compared to the baseline. Thus, the \emph{gained} column
indicates the number of problems newly solved by the ML-enhanced portfolio,
while the \emph{lost} column shows the count of problems that were solved
by the baseline but are not solvable with the ML-enhanced portfolio.

From the baseline improvement, we conclude that \textbf{path features provide a
noticeable improvement of about $\SI{2}\percent$ over the simple bag-of-words features} on
both datasets. Models with \numprint{512} and \numprint{1024} buckets perform
similarly, consistently improving upon the version without path features. In
general, ML strategies do not lose many solutions compared to the baseline, and
path features further enhance this aspect. Interestingly, the best GRUNGE
portfolio with \numprint{1024} buckets does not lose any solutions and even outperforms
the portfolio previously developed for GRUNGE in Table~\ref{tab:ml:grunge}.
Since the lost solutions represent the potential of the baseline strategies to
enrich the ML portfolio, their small values explain why none of the baseline
strategies ranked in the top \numprint{6} of the developed portfolios, despite
being considered during their construction.

\section{Cross-strategy knowledge transfer and model analysis}
\label{sec:exp-extra}

The experiments in Section~\ref{sec:exp} consistently apply strategies using
models trained on their own data. In Section~\ref{sec:cross}, we explore
whether a strategy can benefit from a model trained on data from a different
strategy --- an ability we refer to as \emph{cross-strategy knowledge transfer}.
Section~\ref{sec:exp-analysis} provides insights into the functioning of the
models by conducting a feature importance analysis of the selected models.
Finally, Section~\ref{sec:stats}
concludes with key training and model statistics.

\subsection{Cross-strategy knowledge transfer}
\label{sec:cross}

In the experiments in Section~\ref{sec:exp}, we have consistently paired each
strategy $S$ with a model trained on data generated by $S$. In this section, we
investigate the transferability of knowledge learned from one strategy to
different strategies. Specifically, we examine how well a model trained on data
from one strategy performs when applied to other strategies. We build upon the
experiments from the previous section. In particular, we take the best model
for Mizar and GRUNGE and attempt to use it to guide quantifier selection in
other baseline strategies. Additionally, we merge all the training data for
each dataset and build a combined model, \Mmerge, which we then test with all
baseline strategies.

This experiment evaluates the capability of knowledge transfer between
different strategies. The experiments in this section are conducted using
models with \numprint{1024} buckets. Unlike the previous section, these
experiments are performed on the development set to accelerate evaluation.
Prior results in Section~\ref{sec:exp} have demonstrated that overfitting to
the development set is minimal.
We never mix data from Mizar and GRUNGE, that is, all experiments on the two
datasets are conducted separately.

In the case of Mizar, the model \grk{1} performs best, and therefore, we
evaluate it with all other baseline strategies. Additionally, we assess the
performance of the combined model \Mmerge. The results on Mizar are presented
in Table~\ref{tab:transfer:mizar}.

\begin{table}[h!]
\caption{Knowledge transfer on Mizar/devel.}\label{tab:transfer:mizar}

\begin{center}
\def\arraystretch{1.1}%
\setlength\tabcolsep{0.7em}%
\begin{tabular}{l||r|rr|rr|rr}
    &   \emph{without ML} 
        & \multicolumn{2}{c}{\emph{with ML model S}} &  
           \multicolumn{2}{|c}{\emph{with ML model \sparrow}} &  
           \multicolumn{2}{|c}{\emph{with ML model \Mmerge}} 
\\
\emph{strategy S} & \emph{solves} & \emph{solves} & \emph{gain} [\%] &  \emph{solves} & \emph{gain} [\%] &  \emph{solves} & \emph{gain\%} 
\\\hline
\sparrow     & $\numprint{1454}$ 	& $\mathbf{\numprint{1818}}$ 	& $\mathbf{+25.03\%}$ 	& $\mathbf{\numprint{1818}}$ 	& $\mathbf{+25.03\%}$ 	& $\numprint{1815}$ 	& $+\SI{24.83}\percent$ \\
\chickadee   & $\numprint{1380}$ 	& $\numprint{1731}$ 	& $+\SI{25.43}\percent$ 	& $\numprint{1725}$ 	& $+\SI{25.00}\percent $ 	& $\mathbf{\numprint{1760}}$ 	& $\mathbf{+27.54\%}$ \\
\casc{13}    & $\numprint{1420}$ 	& $\mathbf{\numprint{1677}}$ 	& $\mathbf{+18.10\%}$ 	& $\numprint{1477}$ 	& $+\SI{4.01} \percent $    & $\numprint{1659}$ 	& $+\SI{16.83}\percent$ \\
\buzzard     & $\numprint{1412}$ 	& $\numprint{1718}$ 	& $+\SI{21.67}\percent$ 	& $\numprint{1740}$ 	& $+\SI{23.23}\percent $ 	& $\mathbf{\numprint{1743}}$ 	& $\mathbf{+23.44\%}$ \\
\casc{4}     & $\numprint{755}$ 	& $\numprint{843}$ 	    & $+\SI{11.66}\percent$ 	& $\numprint{851}$   	& $+\SI{12.72}\percent $ 	& $\mathbf{\numprint{858}}$  	& $\mathbf{+13.64\%}$ \\
\casc{16}    & $\numprint{1393}$ 	& $\mathbf{\numprint{1720}}$ 	& $\mathbf{+23.47\%}$ 	& $\numprint{1699}$ 	& $+\SI{21.97}\percent $ 	& $\numprint{1703}$ 	& $+\SI{22.25}\percent$ \\
\hline
\textit{union} & \numprint{1626} & $\mathbf{\numprint{1981}}$ & $\mathbf{+21.83\%}$ & \numprint{1956} & $+20.30\%$ & \numprint{1975} & $+21.46\%$
\end{tabular}

\vspace{2mm}
\small
\begin{tabular}{|ll|} 
    \hline
    \textbf{benchmark:}     & Mizar/devel with \numprint{2896} problems \\
    \textbf{strategies:}    & 6 baseline strategies with different ML models\\
    \textbf{highlighted:}    & best model for each strategy\\
    \hline
\end{tabular}
\normalsize
\end{center}
\end{table}

The column \emph{without ML} presents the individual performance of the
baseline strategies on the development set. The column \emph{with ML model S}
displays the performance of strategy $S$ when paired with a model trained on
its own data. The column \emph{with ML model \grk{1}} shows the performance of
each strategy when guided by model \grk{1}. Finally, the last column,
\emph{with ML model \Mmerge}, reports the performance using the combined model.
The best-performing model for each strategy is highlighted.
The last row \emph{union} presents the cumulative number of problems solved by
the six strategies in each column.

First, all baseline strategies benefit from the ML models. Surprisingly, with
the exception of \casc{13}, there is no significant performance gap between the
models. Nevertheless, better results are obtained with models trained on their
own strategy data and with model \Mmerge rather than with model \grk{1}.
Notably, in the first row for \grk{1}, the first two ML columns correspond to
the same runs, leading to identical results.

For \casc{13}, we observe a significant performance decrease when using model
\grk{1}. Upon inspecting Figure~\ref{fig:strats-casc}, we note that \casc{13}
is the only strategy that employs the option \texttt{pre-skolem-quant=on},
which applies skolemization eagerly to top-level (negatively asserted)
quantified formulas. Since all other strategies without this option benefited
from different models, it is likely that eager skolemization alters the feature
vectors, making the training data incompatible.

The second worst result is observed with \casc{4}, which achieves only half the
relative improvement compared to other strategies. Notably, \casc{4} is the
only baseline strategy that employs finite model finding instead of enumerative
instantiation and e-matching. However, it is remarkable that even a
fundamentally different instantiation approach can still partially benefit from
knowledge learned from related methods.

From the row \emph{total}, we conclude that the best results are obtained when
a strategy is guided by a model trained on its own data. However, a significant
improvement is still achieved when a strategy is guided by a model trained on
data from a different strategy. The total number of problems solved by all \numprint{24}
evaluated strategies is $\numprint{2060}$, indicating a decent level of
complementarity among the strategies.

The results of the same experiments on GRUNGE are presented in
Table~\ref{tab:transfer:grunge}.

\begin{table}[h!]
\caption{Knowledge transfer on GRUNGE/devel.}\label{tab:transfer:grunge}

\begin{center}
\def\arraystretch{1.1}%
\setlength\tabcolsep{0.7em}%
\begin{tabular}{l||r|rr|rr|rr}
    &   \emph{without ML} 
    & \multicolumn{2}{c}{\emph{with ML model S}} &  
    \multicolumn{2}{|c}{\emph{with ML model \casc{1}}} &  
      \multicolumn{2}{|c}{\emph{with ML model \Mmerge}} 
\\
\emph{strategy S} & \emph{solves} & \emph{solves} & \emph{gain} [\%] &  \emph{solves} & \emph{gain} [\%] &  \emph{solves} & \emph{gain} [\%]
\\\hline
\casc{1} 	& $\numprint{473}$ 	& $\mathbf{\numprint{554}}$ 	& $\mathbf{+17.12\%}$ 	& $\mathbf{\numprint{554}}$ 	& $+\mathbf{17.12\%}$ 	& $\numprint{554}$ 	& $+\SI{17.12}\percent$ \\
\casc{6} 	& $\numprint{461}$ 	& $\numprint{503}$ 	& $+\SI{9.11} \percent$ 	& $\numprint{501}$ 	& $+\SI{8.68} \percent $    & $\mathbf{\numprint{509}}$ 	& $\mathbf{+10.41\%}$ \\
\casc{5} 	& $\numprint{441}$ 	& $\mathbf{\numprint{521}}$ 	& $\mathbf{+18.14\%}$ 	& $\numprint{510}$ 	& $+\SI{15.65}\percent $ 	& $\numprint{517}$ 	& $+\SI{17.23}\percent$ \\
\casc{13} 	& $\numprint{448}$ 	& $\mathbf{\numprint{523}}$ 	& $\mathbf{+16.74\%}$ 	& $\numprint{503}$ 	& $+\SI{12.28}\percent $ 	& $\numprint{510}$ 	& $+\SI{13.84}\percent$ \\
\buzzard	& $\numprint{454}$ 	& $\numprint{519}$ 	& $+\SI{14.32}\percent$ 	& $\numprint{515}$ 	& $+\SI{13.44}\percent $ 	& $\mathbf{\numprint{520}}$ 	& $\mathbf{+14.54\%}$ \\
\casc{4} 	& $\numprint{186}$ 	& $\mathbf{\numprint{318}}$ 	& $\mathbf{+70.97\%}$ 	& $\numprint{298}$ 	& $+\SI{60.22}\percent $ 	& $\numprint{311}$ 	& $+\SI{67.20}\percent$ \\
\hline
\textit{union} & $\numprint{527}$ & $\mathbf{\numprint{602}}$ & $\mathbf{+14.23\%}$ & \numprint{592} & $+12.33\%$ & \numprint{591} & $+12.14\%$ 
\end{tabular}

\vspace{2mm}
\small
\begin{tabular}{|ll|} 
    \hline
    \textbf{benchmark:}     & GRUNGE/devel with \numprint{2896} problems \\
    \textbf{strategies:}    & 6 baseline strategies with different ML models\\
    \textbf{highlighted:}    & best model for each strategy\\
    \hline
\end{tabular}
\normalsize
\end{center}
\end{table}

In general, the results on GRUNGE lead to similar conclusions as those on
Mizar. The best results are achieved when strategies are guided by models
trained on their own data. However, even models trained on data from different
strategies provide significant improvements over the baseline. With a total of
612 problems solved by all 24 evaluated strategies, different methods
demonstrate decent complementarity.
Finally, both on Mizar and GRUNGE, merging training data from different
strategies does not lead to a significant gain. While the combined model
\Mmerge performs well, it does not substantially outperform models trained on
individual strategy data. This suggests that strategy-specific models capture
more relevant patterns for guiding their respective strategies than a
generalized model trained on diverse data.

The strategy \casc{13}, which applies eager skolemization, performs worse when
guided by a model trained on \casc{1} data. Although the performance gap is not
as pronounced as in the case of Mizar, it remains noticeable. Interestingly,
the strategy \casc{4}, which employs finite model finding, exhibits a
remarkable individual improvement of more than $70\%$. However, most of the newly
solved problems are already solved by other strategies.

Both on Mizar and GRUNGE, we observe a strong ability for cross-strategy
knowledge transfer, where a model trained on data from one strategy
significantly improves another strategy’s performance. This outcome is not
always guaranteed. For instance, in the context of ATPs, where machine learning
is used to guide \emph{given clause selection}~\cite{JakubuvU18,JakubuvCOP0U20}, such
transfer typically does not occur~\cite{GoertzelJKOPU22}.
A key reason for this difference lies in the syntactic forms of formulas
encountered during proof search. In ATPs, different \emph{term orderings} lead to
distinct normal forms of first-order terms, meaning that strategies operate on
syntactically different formulas, making model transfer challenging. However,
in the context of quantifier selection for SMTs, all strategies process
syntactically equivalent formula representations, which enables effective
knowledge transfer. The performance drop observed for strategy \casc{13}, which
modifies formula structure through eager skolemization, further supports this
claim.

\subsection{Model analysis and feature importance}%
\label{sec:exp-analysis}

We proceed with an analysis of the trained models by examining features present
in the training data and their importance in the model. LightGBM models report
for each feature its \emph{importance}, indicating how much each feature
contributes to the predictions made by the model. %
This aids in understanding which features are most
relevant for the model's performance. 
We collect the training data generated by \grk{1} without path features on the
development set. This strategy was chosen because it is the strongest strategy
without path features on Mizar, and models without path features are simpler
and easier to analyze. We conclude this section with a basic analysis of the
usage of path features.

Without path features, only \numprint{24} bag-of-words features are utilized in
the data, but they exhibit a relatively high number of different values,
especially in the case of context features. 
Many features are present in all training examples. 
Most of the features describe the count of logical
connectives in a formula (\feature{and}{}, \feature{or}{}, \feature{imply}{},
\feature{not}{}, \feature{forall}{}, \feature{equal}{}). 
Additionally, the feature
\feature{uf}{} accumulates the count of first-order (uninterpreted function)
symbols in a formula.
Note that we do not distinguish between different (uninterpreted) symbols names,
and hence our methods are \emph{symbol-independent}~\cite{JakubuvCOP0U20}.
Other features count special symbols, like variables (\feature{var}{}),
bound variables (\feature{bvar}{}), 
quantified variables (\feature{qvar}{}), or
skolem symbols (\feature{skolem}{}).
The key quantifier  feature (marked \textsf{Q}) is symbol count
(\feature{uf}{Q}), while the count of equalities (\feature{equal}{C}) dominates
among context features (\textsf{C}).

\begin{figure*}[t]
    \begin{minipage}{0.625\columnwidth}
   \begin{center}
      \includegraphics[width=1.0\columnwidth]{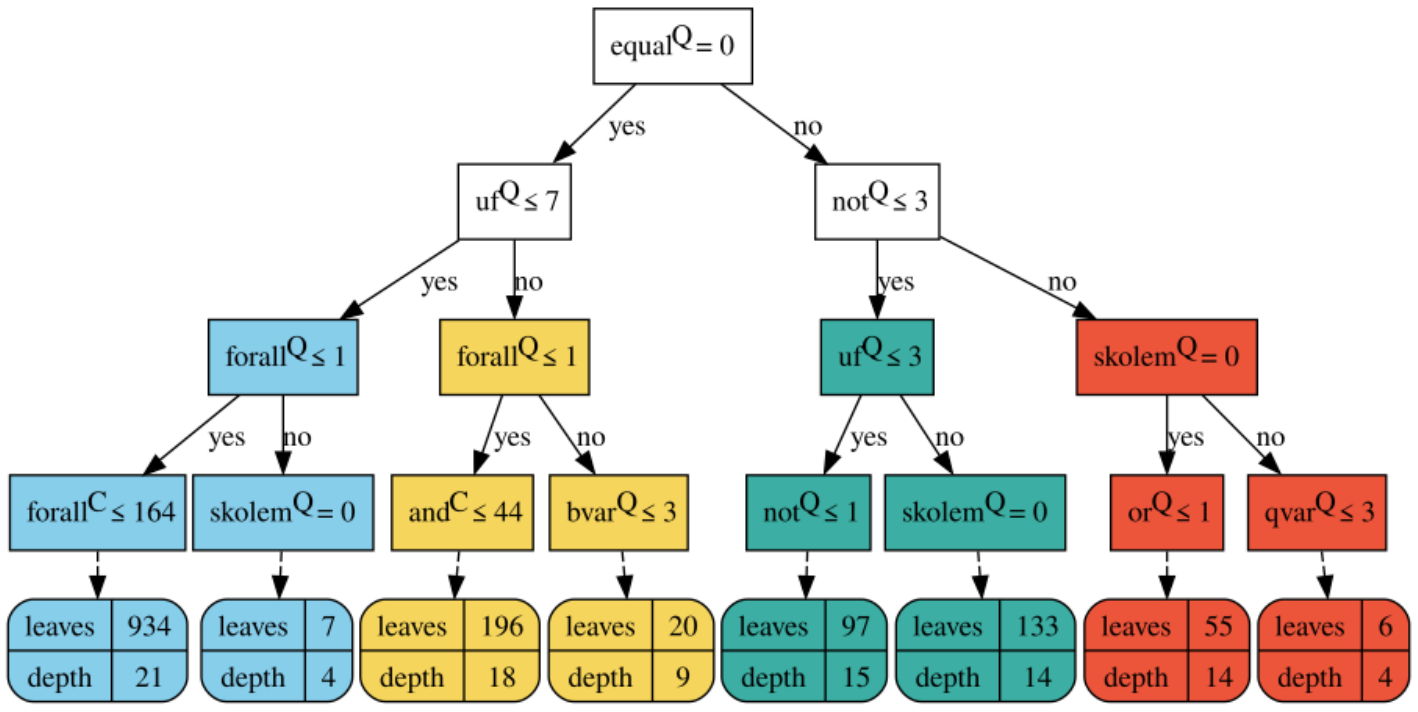}
   \end{center}
\end{minipage}
\begin{minipage}{0.375\columnwidth}
   \begin{center}
      \includegraphics[width=1.0\columnwidth]{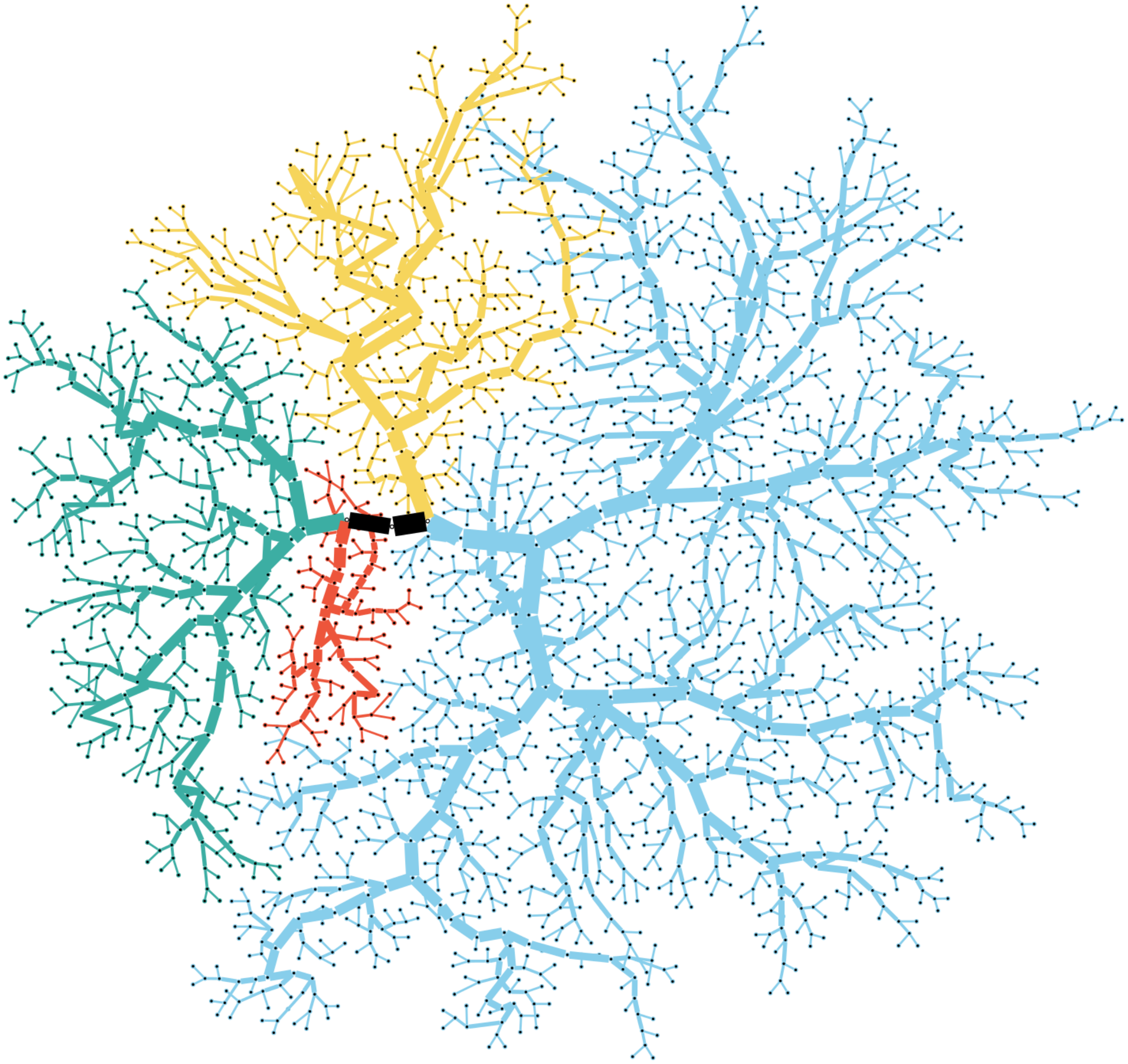}
   \end{center}
   \end{minipage}
\caption{Decision tree visualization.}%
\label{fig:model}
\end{figure*}

Figure~\ref{fig:model} presents a visualization of the first decision
tree in the model \Msparrow. The first of the \numprint{100} trees in the model is
expected to contribute significantly to the final prediction. This decision
tree is a binary tree with inner nodes labeled by conditions on feature values,
while the leaves are labeled by scores. Edges labeled by true or false navigate
each feature vector to a unique leaf to compute the score.  Scores from all
trees are combined into the final prediction.  The root of the first tree in
\grk{1} without path features is located at the left part of Figure~\ref{fig:model}. The root
condition checks the count of equalities in the quantified formula (feature
\feature{equal}{Q}). If the value is \numprint{0}, the ``yes'' edge is followed, and the
next condition $\feature{uf}{Q}\leq 7$ is checked. The first four levels of the
tree are displayed, and the bottom layer indicates the number of leaves in and
the depth of each subtree under that node. 

To provide an overview of the complexity of the tree, the full tree is
visualized on the right part of Figure~\ref{fig:model}. Each edge has a width
proportional to the size of the subtree under this edge. The first two splits
naturally divide the tree into \numprint{4} subtrees, and these subtrees are colored by
corresponding colors in both trees. Therefore, it is apparent that more than
half of the tree (the blue part) is concerned with formulas without equality
($\feature{equal}{Q}=0$) and few symbols ($\feature{uf}{Q}\leq 7$). Note that
context features (like \feature{and}{C}) are used only from the fourth level of
the tree. This tree
visualization illustrates that even with a dozen relatively simple features,
decision trees might be quite complex, with more than \numprint{1500} leaves.

We proceed with the analysis of the best model on Mizar, \grk{1} with
\numprint{1024} buckets (see Table~\ref{tab:ml:mizar}), highlighting the most
frequently used features. The most significant feature is again the
bag-of-words feature \feature{uf}{Q}, which counts the number of uninterpreted
(first-order) symbols in the quantified formula. The second most important
feature is \feature{var}{Q}, which tracks occurrences of bound variables.
The third most important feature is a horizontal feature representing a
unary predicate symbol applied to a variable, such as $P(x)$ in first-order
syntax. Another highly ranked horizontal feature captures a unary function
symbol applied to a variable, like $f(x)$.
The most important vertical path feature corresponds to a negated binary
predicate symbol applied to a variable, such as $\lnot P(x,t_0)$, where
$t_0$ is arbitrary. Since vertical features do not distinguish argument
positions, this could also represent $\lnot P(t_0,x)$.
Other notable path features include the following.
\begin{enumerate}
    \item A horizontal feature counting binary predicates applied to two
        variables (e.g., $P(x,y)$).
    \item A vertical feature capturing a binary predicate applied to a binary
        function with a variable argument (e.g., $P(f(x,t_0),t_1)$).
    \item A vertical feature describing an equality appearing as an argument of
        a disjunction under a universal quantifier, as in $\forall ((t_0=t_1)\lor P_0 \lor P_1)$.
\end{enumerate}

We can conclude that the model effectively utilizes path features to capture
various syntactic patterns in a structured manner. It begins with simpler
patterns, such as unary predicates and functions applied to variables, and
progresses to more complex ones, including binary predicates involving function
applications and quantified expressions with logical connectives. This
highlights the advantage of incorporating path features: while bag-of-words
features offer a high-level statistical summary of formulae, path features
encode structural relationships, enabling the model to make more informed
decisions based on the formula structure.

\subsection{Training and model statistics}%
\label{sec:stats}

We conclude this section with basic statistics of model training, concentrating
on the best model from each dataset. These results also highlight the practical
efficiency of our approach: training requires only standard CPUs, finishes
within minutes, and produces compact models that can be easily stored and
deployed without specialized hardware or large-scale infrastructure.

\begin{description}
    \item [On Mizar,] the best model is \grk{1} with \numprint{1024} buckets.
        It was trained on \numprint{25720} solved training problems, generating
        \numprint{3310297} training vectors, of which $\SI{7.5}\percent$ are
        positive. The size of the text training file is \SI{6}{\giga\byte},
        while the model itself occupies \SI{7.9}{\mega\byte}.

        The model achieves an accuracy of $\SI{88.76}\percent$ on the
        development set, with $\SI{87.83}\percent$ accuracy on positive
        examples. It consists of \numprint{100} trees, each with \numprint{724}
        leaves. Training took approximately \numprint{3} minutes using
        \numprint{64} CPUs.

    \item[On GRUNGE,] the best model is \casc{1} with \numprint{512} buckets.
        It was trained on \numprint{3713} solved training problems, producing
        \numprint{144781} training vectors, of which $\SI{10.15}\percent$ are
        positive. The size of the text training file is \SI{150}{\mega\byte},
        while the model itself occupies \SI{774}{\kilo\byte}.

        The model achieves an accuracy of $\SI{93.43}\percent$ on the
        development set, with $\SI{89.07}\percent$ accuracy on positive
        examples. It consists of \numprint{100} trees, each with \numprint{64}
        leaves. Training took less than a minute using \numprint{64} CPUs.
\end{description}

The number of training vectors generated in our setting is well within the
range commonly considered sufficient for gradient boosting methods such as
LightGBM. For example, the Mizar dataset yields over three million training
vectors, while GRUNGE yields more than one hundred thousand. Comparable or
smaller dataset sizes have been successfully used in related applications of
machine learning to automated reasoning, such as premise selection and proof
guidance~\cite{abs-1108-3446,JakubuvU17}. Gradient boosting in particular is
known to perform robustly with moderate amounts of data, often outperforming
deep learning methods in low- to medium-data regimes~\cite{LightGBM}. 
Thus, our training data can be regarded as sufficient in both quantity and
quality for the task at hand.

Thanks to the short training times, we were able to build multiple models with
varying numbers of leaves and select the best one based on performance on the
development set (see Section~\ref{sec:exp-models}). Notably, the larger size of
the dataset in Mizar results in a higher number of training examples and a
larger model. The most time-consuming task is the generation of training data,
which takes nearly six hours per strategy on Mizar even with heavy
parallelization. Nevertheless, the overall cost of model construction remains
low, demonstrating that our approach is both resource-efficient and practical
for large-scale experimentation without GPUs or distributed clusters.

\section{Interesting Newly Proved Mizar Problems }\label{sec:newprobs}

Our recent work on developing strong strategies for cvc5 has produced
a new state-of-the-art in the number of hard Mizar problems
proved~\cite{cicm24}.  In particular, the total number of the Mizar problems
proved before we started with the work described here was \numprint{46738}, i.e., 80.7\%.
It turns out that the experiments done here strengthen this by another
692 hard Mizar problems, increasing the total number from \numprint{46738} to
\numprint{47430}, i.e., to 81.9\% . In this section we discuss three previously
unproved problems that were proved thanks to the methods developed
here.

The first problem is a topological theorem
URYSOHN3:15\footnote{\url{http://grid01.ciirc.cvut.cz/~mptp/7.13.01_4.181.1147/html/urysohn3.html#T15}}
requiring an 89-line human-written Mizar proof, which is needed for
proving Urysohn's
lemma\footnote{\url{http://grid01.ciirc.cvut.cz/~mptp/7.13.01_4.181.1147/html/urysohn3.html#T20}}~\cite{URYSOHN3}:
a topological space is normal iff two disjoint closed sets can be
separated by a continuous function into the unit interval.

\begin{verbatim}
theorem Th15: :: URYSOHN3:15
for T being non empty normal TopSpace
for A, B being closed Subset of T st A <> {} & A misses B holds
for G being Rain of A,B
for r being Element of DOM
for p being Point of T st (Thunder G) . p < r holds
p in (Tempest G) . r
\end{verbatim}

The problem
has a large number (338) premises because it combines reasoning about
topological spaces, continuous functions, real numbers and dyadic
numbers. This is clausified into 528 clauses that cvc5 has to suitably
instantiate to find the contradiction. The problem's signature
contains 170 symbols, resulting in a fairly complex Herbrand
universe.

The proof is found 
by the \casc{13} strategy, using
an ML model with hashing to 1024 buckets. The proof involves 52
clauses and their 66 ground instances, found after enumerating about
\numprint{7000} instances. This takes about 9 seconds. Since only about 10\% of
the input clauses are needed for the proof, we believe that the
multiple rounds of clause selection implemented by our framework play
an important role here, compared to static premise-selection methods.

The second and third problem are the theorems AFF\_3:1\footnote{\url{http://grid01.ciirc.cvut.cz/~mptp/7.13.01_4.181.1147/html/aff_3.html#T1}} and AFF\_3:2\footnote{\url{http://grid01.ciirc.cvut.cz/~mptp/7.13.01_4.181.1147/html/aff_3.html#T1}} coming from the Mizar article about affine localizations of Desargues's Axiom~\cite{AFF3}. These are two opposite implications:

\begin{verbatim}
theorem :: AFF_3:1
for AP being AffinPlane st AP is satisfying_DES1 holds AP is satisfying_DES1_1

theorem :: AFF_3:2
for AP being AffinPlane st AP is satisfying_DES1_1 holds AP is satisfying_DES1
\end{verbatim}

AFF\_3:1 required an 88-long and AFF\_3:2 a 136-long human-written
Mizar proofs. Unlike in the previous problem, the setting here is much
less diverse and much more uniform: a relatively small geometric
theory. The difficulty lies in dealing with the relatively large
configurations of points and lines that allow many many possible
instantiations of the available premises. E.g., for AFF\_3:2, the
proof requires 97 ground instances, and a single premise has to be
suitably instantiated up to 20 times here. Also the formulas and
clauses often contain many similar predicates and literals, which
typically make efficient indexing and subsumption hard and result in a
quick combinatorial explosion in standard saturation-style ATPs. Here
we give two examples of such combinatorial formulas used in the proofs:

\begin{verbatim}
theorem Th6: :: AFF_1:6
for AS being AffinSpace
for x, y, z being Element of AS st LIN x,y,z holds
LIN x,z,y & LIN y,x,z & LIN y,z,x & LIN z,x,y & LIN z,y,x 

theorem Th7: :: AFF_1:7
for AS being AffinSpace
for x, y being Element of AS holds LIN x,x,y & LIN x,y,y & LIN x,y,x 
\end{verbatim}

The proof search here used our \sparrow{} strategy and an ML model with
hashing to \numprint{1024} buckets.
The search generated almost \numprint{60000} instances 
and took 40 seconds.

\section{Conclusions and future work}\label{sec:conclusion}

This paper develops a novel approach that trains a machine learning model to decide
which quantifiers should be instantiated inside an SMT solver.
The approach is conceptually simple: in each instantiation round, each
quantifier is considered to be selected for instantiation or not; the exact way
in which the quantifier is instantiated is left to the SMT solver.
This enables us to employ our approach for various instantiation
techniques~\cite{DetlefsNS05,reynolds-cade13,janota2021fair}.
This contrasts with existing research that always instantiates \emph{all}
active quantifiers and tries to modify \emph{one specific instantiation
strategy} with an ML model~\cite{janota-sat22,ourao} --- and it is limited to
that particular instantiation strategy. Also, in contrast to the existing work,
we showed that we can boost the solver's performance if we take a portfolio of
its configurations. This is a tall order. Indeed, it is highly challenging to
improve one particular configuration of the solver in a way that it is not
covered by another configuration of the solver. 
Thanks to the versatility of our approach, we could integrate it with various
instantiation strategies, thereby enhancing multiple solver configurations and
achieving a notable \textbf{$\SI{22.46}\percent$ improvement over the baseline
state-of-the-art portfolio on Mizar and $\SI{12.86}\percent$ improvement on GRUNGE}.
The key aspects contributing to this success include our ability to implement
our selection method with \textbf{different instantiation modules} and effectively
\textbf{transfer learned knowledge} across strategies. 
Further improvement is achieved by extended \textbf{path features}.
This enables us to substantially
enhance each individual strategy within the portfolio, consequently boosting
the overall portfolio performance.
Our best strategy solves \numprint{1845} in \numprint{60} seconds while it solves
\numprint{1810} in \numprint{30} seconds.
This number can be informatively compared with experiments on the same dataset
from the literature~\cite{GoertzelCJOU21}, where the state-of-the-art ATP
prover E with advanced machine learning methods, solves only \numprint{1632}
of the holdout problems in \numprint{30} seconds. 

Our approach differs from existing research that aimed to learn which
quantifiers to select on a single problem instance~\cite{bandits} --- using the
multi-armed bandit paradigm (MAB). While previous attempts failed to achieve
experimental improvements, combining MAB with offline training presents an
intriguing avenue for future exploration.

We have extended our previous methods and results on Mizar~\cite{JakubuvJPU24}
with enhanced formula features and we have successfully performed experiments
on the GRUNGE dataset.
On GRUNGE, we were able to train successful models from significantly smaller
amount of training data.
Another logical next research direction is to apply and assess our methodology on
various benchmarks. We conducted evaluations on Mizar MML problems due to their
similarity and relevance, enhancing the potential for machine learning methods
to recognize useful quantified formulas.

More diverse benchmarks like TPTP~\cite{SutcliffeSY94} or
SMT-LIB~\cite{BarST-SMT-10} still present significantly greater challenges. These
benchmarks contain problems from various sources, often with few related
instances, limiting learning opportunities. Furthermore, the transfer of
learned knowledge across different SMT logics complicates the situation with
SMT-LIB\@. 

\bibliographystyle{elsarticle-num}
\bibliography{all}

\end{document}